\documentclass{article} % For LaTeX2e
\usepackage{fullpage}
\usepackage[utf8]{inputenc} % allow utf-8 input
\usepackage[T1]{fontenc}    % use 8-bit T1 fonts
\usepackage{url}            % simple URL typesetting
\usepackage{booktabs}       % professional-quality tables
\usepackage{amsfonts}       % blackboard math symbols
\usepackage{nicefrac}       % compact symbols for 1/2, etc.
\usepackage{microtype}      % microtypography
\usepackage{soul}
\usepackage{graphicx}
\usepackage{amsmath}
\usepackage{outlines}
\usepackage{xurl}

\usepackage[colorlinks=true,
citecolor=blue,
filecolor=black,
linkcolor=blue,
urlcolor=blue]{hyperref}

\usepackage{amsmath,amsfonts,bm}

\def\eqref#1{Equation~\ref{#1}}
\def\1{\bm{1}}

\def\rvx{{\mathbf{x}}}

\DeclareMathAlphabet{\mathsfit}{\encodingdefault}{\sfdefault}{m}{sl}
\SetMathAlphabet{\mathsfit}{bold}{\encodingdefault}{\sfdefault}{bx}{n}

\usepackage{multirow}
\usepackage{paralist}

\usepackage{multicol}
\usepackage{diagbox}

\usepackage[ruled,noend]{algorithm2e}

\SetCommentSty{mycommfont}

\usepackage{here}

\usepackage{amsmath,amssymb,amsfonts,amsbsy,amsfonts,latexsym}
\usepackage{makecell}
\usepackage{xcolor}
\usepackage{colortbl}

\usepackage{tabularx,colortbl,xcolor}
\usepackage[normalem]{ulem}
\useunder{\uline}{\ul}{}

\usepackage{enumitem}

\usepackage{xparse}% http://ctan.org/pkg/xparse

\SetKwInput{KwInput}{Input}
\SetKwInput{KwRequire}{Require}

\usepackage{longtable}

\NewDocumentCommand{\var}{O{s} m O{}}{%
  \ensuremath{#1_{#2}^{#3}}% add \vphantom{<bizarre sup>}
}
\usepackage{siunitx}

\newcommand{\commentout}[1]{}

\definecolor{light-gray}{gray}{0.80}

\newcommand\fref{Figure~\ref}
\newcommand\tref{Table~\ref}
\newcommand\sref{Section~\ref}

\newcommand\hb{ \rowcolor{orange!15}}
\newcommand\hc{ \rowcolor{orange!40}}

\newcommand\gb{ \rowcolor{gray!15}}
\newcommand\gc{ \rowcolor{gray!40}}

\usepackage{amsthm}

\usepackage{amsthm}

\newtheorem{mytheorem}{Theorem}
\newtheorem{assumption}[mytheorem]{Assumption}
\newtheorem{lemma}[mytheorem]{Lemma}
\newtheorem{remk}[mytheorem]{Remark}

\newcommand{\OURS}{ZeroQuant(4+2)\xspace}

\usepackage{xcolor}
\usepackage{listings}
\makeatletter
\def\adl@drawiv#1#2#3{%
        \hskip.5\tabcolsep
        \xleaders#3{#2.5\@tempdimb #1{1}#2.5\@tempdimb}%
                #2\z@ plus1fil minus1fil\relax
        \hskip.5\tabcolsep}
\newcommand{\cdashlinelr}[1]{%
  \noalign{\vskip\aboverulesep
           \global\let\@dashdrawstore\adl@draw
           \global\let\adl@draw\adl@drawiv}
  \cdashline{#1}
  \noalign{\global\let\adl@draw\@dashdrawstore
           \vskip\belowrulesep}}
\makeatother

\definecolor{upforestgreen}{rgb}{0.6, 0.8, 0.2}

\usepackage{subfig}

\usepackage{chngcntr}
\usepackage{adjustbox}

\usepackage{wrapfig}
\usepackage{arydshln}

\usepackage{tcolorbox}

\usepackage{pifont}% http://ctan.org/pkg/pifont
\newcommand{\cmark}{\ding{51}}%
\newcommand{\xmark}{\ding{55}}%

\newcommand{\ptq}{PTQ\xspace}

\newcommand{\codellama}{CodeLLaMA\xspace}
\newcommand{\codestar}{StarCoder\xspace}
\newcommand{\codegee}{CodeGeeX2\xspace}

\newcommand{\bart}{BART$_{\text{406M}}$\xspace}
\newcommand{\llama}{LLaMA\xspace}
\newcommand{\rtn}{RTN\xspace}

\newcommand{\ourformat}{FP6$_{E3M2}$\xspace}

\begin{document}

% \title{: A New High-Performance FP6-Centric Quantization Strategy for LLMs}
% }
\title{\OURS: Redefining LLMs Quantization with a New FP6-Centric Strategy for Diverse Generative Tasks}

\author{Xiaoxia Wu\thanks{Our corresponding authors are Xiaoxia Wu (xiaoxiawu@microsoft.com) and Leon Song (leonsong@microsoft.com). }, Haojun Xia\thanks{Haojun Xia is currently a PhD student at The University of Sydney, Australia.}, Stephen Youn, Zhen Zheng, Shiyang Chen\thanks{Shiyang Chen is currently a PhD student at Rutgers University, USA.}, 
Arash Bakhtiari,\\ Michael Wyatt, Reza Yazdani Aminabadi, Yuxiong He, Olatunji Ruwase,\\ Leon Song$^*$, Zhewei Yao
\\  \\ DeepSpeed of Microsoft}
\date{}

\maketitle

%%%%%%%% BODY TEXT
\begin{abstract}
% This study evaluates 4-bit quantization methods like GPTQ in large language models (LLMs) and reveals GPTQ's tendency of overfitting with limited Zero-Shot performance enhancement. While prior works merely focusing on zero-shot measuremean, we extend the tasks to more generative task such as code generation and summarizaiton, in which we observe INT4 quantization's underperformance in generative tasks. Thus, we shift our focus to the higher precision formats: FP6. Previously, these formats were overlooked due to the lack of sophisticated system support and acceleration strategies on modern GPUs. In this work, we propose a novel 4+2 algorithmic design to facilitate FP6's large-scale integration and adaptation in the LLM era. Our findings demonstrate FP6's with coarse-grain quantization scheme already robust performance across a spectrum of quantization algorithms, ensuring its accuracy and versatility. Particularly, we find the FP-6 \llama-34B perform almost the same as it's counterpart in code generation, even smaller models 0.5B are the same as it's baseline. We position FP6 as a viable alternative for tackling the limitations of current 4-bit quantization methods for LLMs.

This study examines 4-bit quantization methods like GPTQ in large language models (LLMs), highlighting GPTQ's overfitting and limited enhancement in Zero-Shot tasks. While prior works merely focusing on zero-shot measurement, we extend task scope to more generative categories such as code generation and abstractive summarization, in which we found that INT4 quantization can significantly underperform. However, simply shifting to higher precision formats like FP6 has been particularly challenging, thus overlooked, due to poor performance caused by the lack of sophisticated integration and system acceleration strategies on current AI hardware. Our results show that FP6, even with a coarse-grain quantization scheme, performs robustly across various algorithms and tasks, demonstrating its superiority in accuracy and versatility. Notably, with the FP6 quantization, \codestar-15B model performs comparably to its FP16 counterpart in code generation, and for smaller models like the 406M it closely matches their baselines in summarization. Neither can be achieved by INT4. To better accommodate various AI hardware and achieve the best system performance, we propose a novel 4+2 design for FP6 to achieve similar latency to the state-of-the-art INT4 fine-grain quantization. With our design, FP6 can become a promising solution to the current 4-bit quantization methods used in LLMs.\footnote{Code will be released soon as a part of \url{https://github.com/microsoft/DeepSpeed}}

\end{abstract}
% Inspired by prior works on data pruning and layer sensitivity studies on multi-head attention, 

\section{Introduction}

Large Language Models (LLMs) such as GPT-3 \cite{brown2020language} have significantly advanced the field of natural language processing. These models have shown exceptional capabilities in various complex tasks, from text generation to language understanding. However, the widespread adoption of LLMs is challenged by their extensive computational and memory demands. This issue is particularly acute in resource-constrained environments, where deploying such large models is not feasible. To mitigate these challenges, post-training quantization has been recognized as a crucial technique \cite{cai2020zeroq,gholami2021survey,nagel2020up,liu2021post}. It enables the compression of these models for efficient utilization in limited-resource settings without the need for extensive retraining. Nevertheless, this approach often necessitates a balance between reducing the model size and maintaining accuracy \cite{dettmers2022case}.

Recent developments in the field of quantization, particularly in 4-bit quantization, have demonstrated potential in compressing LLMs effectively as their quality drops are greatly minimized due to advance algorithm design such as GPTQ~\cite{frantar2022gptq} and LoRC~\cite{yao2023zeroquant}. However, these advancements have predominantly focused on zero-shot evaluations and the acceptable quality drops are for larger model size greater 13B, yet they often come with a significant trade-off for smaller model size such as $1$B. Moreover, they only focus on zero-shot measurement~\cite{wu2023zeroquant, yao2023zeroquant}. In production environments, where replicating the original model's performance across different tasks is critical, any loss of model quality is a major concern. Existing methods, while innovative, do not fully address the practical requirements for deploying LLMs in real-world applications.

% \textbf{Contribution.} In response to these limitations, we make the following contribution:
% \begin{itemize}
% \item \textbf{Enhanced Task Scope and Quantization Analysis.} We find that the current quantizatio method like GPTQ overfitted to calibrated datsets. More imporatnatly we expanded the examination of 4-bit quantization methods in LLMs beyond Zero-Shot tasks, including code generation and abstractive summarization and find the INT4 can siginicitantly underperform particularly for small model even like 13b such as \llama-13b.
% \item \textb{Robust Performance of FP6 Quantization.} We  assure that  FP6, even with coarse grain quantization  stands out for its ability to maintain accuracy levels comparable to full-precision models, making it versatile for a range of applications. Notably, with the FP6 quantization, \codellama-34B model performs comparably to its FP16 counterpart in code generation, and for smaller models like the 406M it closely matches their baselines in summarization. Neither can be achieved by INT4. 
% \item More importantly, Introduced a novel 4+2 design for FP6, addressing previous integration and acceleration challenges on AI hardware. This design achieves similar latency to state-of-the-art INT4 fine-grain quantization, positioning FP6 as a promising alternative for current 4-bit methods in LLMs. 
% \end{itemize}

\textbf{Contribution.} To address these challenges, our contributions are as follows:
\begin{itemize}
    \item \textbf{Broadened Evaluation Scope and Quantization Analysis.} Our study reveals that existing quantization methods like GPTQ tend to overfit to calibrated datasets. More significantly, we have broadened the scope of 4-bit quantization analysis in LLMs to include tasks beyond Zero-Shot, such as code generation and abstractive summarization. We discover that INT4 quantization often underperforms, especially in smaller models, even as large as 13 billion parameters, exemplified by \llama-13b. See \sref{sec:int4} for details.
    
    \item \textbf{Superior Performance with FP6 Quantization.} 
We illustrate that FP6, employing a basic round-to-nearest (RTN) algorithm and a coarse-grain quantization approach, consistently achieves accuracy on par with full-precision models, proving highly effective across a broad spectrum of generative tasks. The \codestar-13B model with FP6 quantization matches the performance of its FP16 equivalent in code generation tasks. For smaller models such as the 406M, it aligns closely with baseline results in summarization. These achievements are beyond the capabilities of INT4 quantization. For a more in-depth exploration, refer to Section \ref{sec:methodology}.
    
    \item \textbf{Innovative 4+2 FP6 Design.} We introduce an innovative 4+2 design for FP6 that overcomes prior integration and acceleration issues on AI hardware. This design attains latency similar to the state-of-the-art INT4 fine-grain quantization, establishing FP6 as a viable alternative to existing 4-bit quantization methods in LLMs. See \sref{sec:system} for details.
\end{itemize}

\section{Related Work}\label{sec:related_work}
In this study, we specifically focus on the quantization of Large Language Models (LLMs), diverging from other neural network architectures like BERT and ResNet models, which have been extensively explored in existing literature~\cite{shen2020q,zafrir2019q8bert,fan2020training,wu2022extreme,bai2020binarybert,esser2019learned,tao2022compression,kim2021bert}. 

Quantization generally refers to employing low-precision weights and activations to leverage faster arithmetic cores, such as INT8/INT4 tensor cores~\cite{hubara2017quantized}. However, the distinctive bandwidth constraints of LLMs have popularized weight-only quantization methods~\cite{zafrir2019q8bert,frantar2022gptq, yao2022zeroquant,wu2023zeroquant,wu2023understanding} as a strategy to reduce the memory footprint of these models.

Most previous research evaluates the impact of quantization using metrics like zero-shot perplexity or accuracy~\cite{xiao2022smoothquant,frantar2022gptq,chee2023quip,ashkboos2023towards,kim2023squeezellm}. However, given that the main real-world applications of LLMs, such as ChatGPT~\cite{brown2020language} and Codex~\cite{copilot}, revolve around generation-based tasks, a more comprehensive evaluation framework for quantized LLMs is warranted.

While many studies focus on integer data formats for their ease of simulation and extensive ecosystem support~\cite{krishnamoorthi2018quantizing,dong2019hawq,frantar2022gptq,chee2023quip,kim2023squeezellm,kim2023memory}, recent works have also demonstrated the effectiveness of floating-point formats~\cite{wu2023zeroquant,dettmers2023qlora}. Nonetheless, these investigations typically center on conventional bit precisions such as 2/4/8 bits. Some research, like GPTQ, delves into 3-bit precision, but number concatenation methods, as discussed in Section~\ref{sec:system}, limit their system performance.

Finally, while the push for lower precision quantization continues, the practicality of deploying a model of size \( xB \) quantized to \( 2y \) bits over a \( 2xB \) model with \( y \) bits quantization is often overlooked, despite potential quality advantages~\cite{dettmers2022case,yao2023zeroquant}. Our paper seeks to find an optimal balance where the quantized model retains similar accuracy to a full-precision model, an aspect largely missing in current literature~\cite{park2022nuqmm, yao2023zeroquant, shang2023pb,yuan2023rptq,lee2023owq,guo2023lq,wei2023outlier,guo2023olive,yao2023zeroquant-hero}.

\section{Comprehensive Evaluation is Needed}\label{sec:int4}
For completeness, we here explain some foundational terminology and concepts in quantization. 

\textbf{Integer Quantization.} Consider a full-precision $\rvx\in \mathbb{R}^d$ and its quantized counterpart $\rvx_{q}\in \mathbb{R}^d$. The  mapping strategy from $\rvx$ to $\rvx_{q}$ is
\begin{equation}
    \rvx_{q} = S\left\lceil clamp( (\rvx-x_{\text{zero}}\mathbf{1})/S; 0, 2^{b-1}-1)\right\rceil +x_{\text{zero}} \mathbf{1},\label{eq:quant}
\end{equation}
where $clamp$ the function map its input values to  a given range from $-2^{b-1}$ to $2^{b-1}-1$. $x_{\text{zero}}$ serves as a reference point, aiming to minimize any bias and can be defined as the minimum value, such as $x_{\text{zero}}=min(\rvx)$. 
%a lower bound of $-2^{b-1}$ and an upper bound of $2^{b-1}-1$  to the value of $\rvx/S$ --- from $-2^{b-1}$ to $2^{b-1}-1$
 The parameter $b$ represents  the number of bits used to represent the quantized value. $\lceil\cdot\rceil$ is the rounding operator, and $S \in \mathbb{R}$ is the scaling factor. 
For example, $S$ can be  computed as the distance between the maximum and minimum of the absolute elements in the  $\rvx$ vector, i.e., $S = max(\rvx)-min(\rvx)$.  For 4-bit integer (INT4) quantization on LLMs, this algorithm with $x_{\text{zero}}$ strictly greater zero (called asymmetric quantization) has been widely adopted and proved to be much better than that by setting $x_{\text{zero}}=0$ based on prior works ~\cite{wu2023understanding,wu2023zeroquant,yao2022zeroquant,yao2023zeroquant}. For history and details on how to set the parameters, see~\cite{gholami2021survey}.

% \textbf{Fine-grain Quantization (FGQ) and coarse-grain Quantization (CGQ).}  has emerged as a leading approach in the quantization of LLMs. It involves adjusting the precision at a more detailed level within each row or column of a matrix, aiming to either maintain or enhance the model's performance (such as accuracy and generation quality) while also realizing efficiency gains. 

\textbf{Fine-grain Quantization (FGQ) and Coarse-grain Quantization (CGQ)} relates to the value of $d$ in previous paragraph. Suppose a matrix  of dimension \(n \times n\) is vectorized into \(n^2\) elements which are then grouped into blocks of size \(d\), yielding \({n^2}/{d}\) groups. When the block size \(d\) is set to 1, it equates to the matrix's original precision. FGQ comes into play when the block size is smaller than specific thresholds like 64, 128, or 256, given a matrix size of \(n \geq 1024\). This method focuses on adjusting the precision within each row or column of the matrix at a finer level. CGQ, in contrast, uses a block size equivalent to the row size (\(d \geq 1024\) in this case), resulting in a coarser quantization approach. FGQ gained significant attention in the realm of LLMs because the values in the weight matrices tend to have a wider distribution, as noted in~\cite{yao2023zeroquant}.

Alongside FGQ or CGQ, specific algorithms are employed for precision mapping in quantization. Given the focus on 4-bit quantization and the demonstrated efficacy of the INT4 format over FP4 (as detailed in the appendix) \cite{wu2023zeroquant}, the investigation primarily centers on a straightforward method, \rtn, and the increasingly recognized and impactful algorithm, GPTQ~\cite{frantar2022optimal,frantar2022gptq}, with a solid foundation background~\cite{lecun1990optimal,hassibi1993second}. We now explain them briefly below:

\begin{itemize}
\item 
\noindent\textbf{\rtn.} \textbf{R}ound-\textbf{t}o-\textbf{n}earest neighborhood simply map the weight matrices to its low-precision counterpart based on \eqref{eq:quant}. %Specifically, for symmetric quantization, we set $S=max(abs(x))$ and $Z=0$; for asymmetric quantization, we set $S=max(x)-min(x)$ and $Z=min(x)$.and follows the procedure detailed in Section~\ref{sec:background_of_quantization}

\item\noindent\textbf{GPTQ.} \textbf{G}enerative \textbf{P}re-trained \textbf{T}ransformer \textbf{Q}uantization is a more advanced method of leveraging the activation information, which requires the inverse of the second-order input information.   According to \cite{frantar2022optimal,frantar2022gptq}, it reduces the precision of the model's weights to a lower bit representation (down to 3 or 4 bits per weight) without significant accuracy loss. Their code implementation is structured in a layer-by-layer manner, transferring the computational burden to the CPU when it's not in use. This strategy allows for the execution of massive models, like those with 175B parameters, on a single GPU, overcoming previous limitations of scale and complexity. GPTQ enhances the practical deployment of these models, particularly in memory and computationally constrained environments.

\end{itemize}
\begin{table}[tb]
 \centering
  \caption{\textbf{Zero-Shot Evaluation (Perplexity$\downarrow$)}. GPTQ quantization algorihtm for INT4 weight (W4A16) on \llama-1B (Left) and \llama-13B (Right). Different calibration datasets result in different results.}\label{table:ppl-int4}
 \begin{adjustbox}{width=0.899\linewidth}
 \begin{tabular}{lcc|cccc|ccccccc}\toprule
             & & Dataset &\multicolumn{4}{c}{\llama-1B (4096-seq)}     &  \multicolumn{4}{|c}{\llama-13B (2048-seq)}      \\
       Precision & FGQ & for GPTQ & PTB   & PTB-new & C4  & C4-new &  PTB   & PTB-new & C4  & C4-new\\\midrule 
FP16   & N/A   & N/A     & 37.39  & 58.34   & 8.91 & 9.4    & 19.23 & 28.10    & 6.61 & 6.8   \\ \cdashline{1-12}
\multirow{3}{*}{INT4-GPTQ} &  \xmark & PTB   & \textbf{49.80}   & \textbf{64.01}   & 10.00   & 10.49 & \textbf{19.68} & \textbf{28.71}   & 6.91 & 7.17   \\
  & \xmark&  C4    & 719.21 & 693.48  & {9.84} & {10.37} & 21.31 & 30.01   & {6.84} & {7.09} \\ %\cdashline{1-10}
% \multirow{2}{*}{INT4-GPTQ} & PTB &  \textbf{60.05}&  \textbf{69.40} &9.37& \textbf{9.87} & \textbf{19.37}&28.10&6.77&6.98\\
   & \cmark  &  C4& 1399.89&1396.76&\textbf{9.34}&\textbf{9.84} & 22.14&29.83&\textbf{6.74}&\textbf{6.95} \\ 
\bottomrule  
\end{tabular}
\end{adjustbox}
\end{table}

\begin{table}[tb]
 \centering
 \caption{\textbf{Zero-Shot Evaluation (Perplexity$\downarrow$).} Compare between GPTQ$_{C4}$ and RTN quantization algorithms for INT4 weight (W4A16) on \llama of size 1B, 13B and 65B. We apply  fine-grain quantization (FGQ) in which the block-size is 256 elements per scale (\llama-1B's block-size is 128). We also report results for coarse-grain quantization (CGQ) (per row per scale). The evaluation datasets are  Wikitext2, PTB, C4, PTB-new, and C4-new.}\label{table:pp-int4a}
 \begin{adjustbox}{width=0.999\linewidth}
\begin{tabular}{lllll}\toprule
 Quant   & Precision & \llama-1B (4096-seq)                                          & \llama-13B (2048-seq)                                     & \llama-65B (2048-seq)                                  \\\midrule
    & FP16       & 24.31$_{7.53/37.39/8.91/58.34/9.40 }$      & 13.16$_{5.09/19.23/6.61/28.10/6.80 }$ & 6.41$_{3.56/8.00/5.62/8.88/5.98 }$  \\\midrule
FGQ & INT4-GPTQ  & 564.73$_{7.83/1399.89/9.34/1396.76/9.84 }$ & 14.19$_{5.28/22.14/6.74/29.83/6.95}$  & 6.61$_{3.81/8.17/5.73/9.20/6.13 }$  \\
    & INT4-RTN   & 22.76$_{7.98/34.03/9.50/52.28/10.00 }$     & 14.32$_{5.35/22.49/6.80/29.96/7.00 }$ & 7.30$_{3.79/10.13/5.74/10.54/6.31}$  \\\midrule
CGQ & INT4-GPTQ  & 288.22$_{8.20/719.21/9.84/693.48/10.37 }$  & 14.13$_{5.37/21.31/6.84/30.01/7.09 }$ & 7.17$_{4.12/10.50/5.83/9.16/6.21 }$ \\
    & INT4-RTN   & 34.29$_{8.33/55.52/10.15/86.85/10.62 }$    & 14.32$_{5.55/20.95/6.97/30.91/7.22 }$ & 7.72$_{4.20/10.59/5.90/11.44/6.47}$\\
\bottomrule  
\end{tabular}
\end{adjustbox}
\end{table}

\begin{table}[tb]
 \caption{\textbf{Zero-Shot Evaluation (Accuracy$\uparrow$)}. Compare between GPTQ$_{C4}$ and RTN quantization algorithms for INT4 weight (W4A16) on \llama-1B (Top) and \llama-13B (Bottom). We apply fine-grain quantization (FGQ) in which the block-size is 256 elements per scale except for \llama-1B's (which is 128). arcC (arcE) stands for arc\_challenges (arc\_easy).
 }\label{table:zero-shot-int4}
 \begin{adjustbox}{width=0.999\linewidth}
 \begin{tabular}{lllllllllllll}\toprule
        Models             &   Precision (FGQ)    &  arcC & arcE & boolq & cb    & copa  & piqa  & rte   & wic   & wsc   & storycloze & MEAN \\\midrule 
\multirow{2}{*}{\llama-1B }&  FP16 & 26.71 & 53.11 & 61.13 & 39.29 & 76.00    & 73.83 & 50.18 & 50.00    & 36.54 & 69.64 & 53.64   \\\cdashline{2-13}
\multirow{2}{*}{(4096-seq)}&  INT4-GPTQ    & \textbf{26.37}          & 50.59     &  \textbf{61.59} & 46.43 &  \textbf{79.00}    & \textbf{73.34} & 48.01 & 50.00    & 36.54 & 68.24      & \textbf{54.01} \\
                     &  INT4-RTN & 26.11          &  \textbf{51.09}     & 58.07 &  \textbf{50.00}    & 74.00    & 72.91 & \textbf{48.38} & 50.00    & 36.54 & \textbf{68.36}      & 53.55 \\ \midrule
\multirow{2}{*}{\llama-13B}& FP16 & 43.86 & 74.58 & 68.53 & 50.00 & 90.00 & 79.00 & 65.34 & 50.00 & 35.58 & 78.23 & 63.51   \\\cdashline{2-13}
\multirow{2}{*}{(2048-seq)}&  INT4-GPTQ    & 43.00          & 73.44     & \textbf{67.83} & 41.07 & \textbf{93.00} & 78.78 & 62.45 & \textbf{50.16} & 36.54 & 78.17      & 62.44 \\
                     & INT4-RTN & \textbf{44.03}          & \textbf{74.45}     & 67.37 & \textbf{44.64} & 91.00 & \textbf{78.84} & \textbf{63.18} & 49.84 & 36.54 & \textbf{78.42}      & \textbf{62.83}\\\midrule
    \multirow{2}{*}{\llama-65B}& FP16 &  47.01 & 75.08 & 82.32 & 64.29 & 91.00 & 81.61 & 71.48 & 58.31 & 60.58 & 79.57 & 71.13  \\\cdashline{2-13}
\multirow{2}{*}{(2048-seq)}&  INT4-GPTQ    &46.84 & 75.08 & 80.76 & 58.93 & 94.00 & 81.18 &  \textbf{72.92} & 56.27 & 60.58 & 79.31 & 70.59  \\
                     & INT4-RTN   & \textbf{47.10} &  \textbf{75.25} &  \textbf{81.47} &  \textbf{62.50} &  \textbf{95.00} &  \textbf{81.23} & 69.68 &  \textbf{57.21} &  \textbf{62.50} &  \textbf{79.63} & \textbf{71.16} \\
%                      \midrule
% \multirow{2}{*}{\llama-65B}& FP16 &   \\\cdashline{2-13}
% \multirow{2}{*}{(2048-seq)}&  INT4-GPTQ    &  \\
%                      & INT4-RTN & \\
\bottomrule  
\end{tabular}
\end{adjustbox}
%\end{table}
\\ 
\vspace{0.3cm}
\\
%\begin{table}[tb]
\caption{\textbf{Generation Tasks (Rouge$\uparrow$ or Pass@1$\uparrow$)}. INT4 weight (W4A16) quantization on \bart, \codegee-6B, \codestar-15B, and \codellama-34B models (left to right) Using RTN. Note that summarization tasks use two separate \bart versions fine-tuned by CNN/XSUM and code generation tasks in Human-X including Python and JavaScript (as the variances of other tasks such as CPP, Go and RUST are higher and so not included), averaged over at least 8 repeated experiments with standard deviation.} \label{table:gen-int4a}
\begin{adjustbox}{width=0.999\linewidth}
\begin{tabular}{l|c|cc|cc|cc}\toprule 
             & \multicolumn{1}{c|}{\bart {\small (R1/R2/RL)}}    & \multicolumn{2}{c|}{\codegee-6B {\small (pass@1)}}               & \multicolumn{2}{c|}{\codestar-15B {\small(pass@1)}}    & \multicolumn{2}{c}{\codellama-34B {\small(pass@1)}}             \\
   Precision    &      XSUM (CNN)  & Python        & Java-Scrpit       & Python        & Java-Scrpit  & Python        & Java-Scrpit          \\\midrule 
%  FP16      & 44.07/21.09/30.65/41.02      & 45.49/22.39/37.28/37.28 & 34.04$\pm$1.70 & 31.50$\pm$2.62  & 35.43$\pm$2.21  & 33.67$\pm$2.02 \\
% INT4-RTN      &  43.67/20.68/30.07/40.62      & 43.76/20.76/35.82/35.82 & 33.08$\pm$2.07  & 25.15$\pm$1.97 & 33.20$\pm$1.40  & 32.18$\pm$1.29  \\
% 2 & \ourformat & 34.10$\pm$2.09  & 27.74$\pm$1.18 & 31.61$\pm$1.74 & 31.15 & 35.64$\pm$1.26 & 27.91$\pm$1.49 & 33.60$\pm$1.91  & 32.38 \\

  FP16      &   45.49/22.39/37.28 (44.07/21.09/30.65)  & 34.04±1.70  & 31.50±2.62  & 35.43±2.21 & 33.67±2.02 & 43.22±2.21 & 45.05±1.60  \\
% 44.06/21.07/30.669/40.9938  & 45.37/22.2/37.1095/37.1087   & 34.1±2.09  & 31.61±1.74 & 35.64±1.26 & 33.6±1.91  & 44.31±1.88 & 44.51±1.3  \\
INT4 (CGQ)     & 43.76/20.76/35.82 (43.67/20.68/30.07) & 33.08±2.07 & 25.15±1.97 & 33.20±1.40   & 32.18±1.29 & 39.84±1.64 & 43.45±2.05 \\
INT4 (FGQ)  & 44.82/21.63/36.48 (43.73/20.72/30.24)   &29.80$\pm$1.76 &28.35$\pm$2.36 & 35.64$\pm$2.52 & 32.32$\pm$2.01  & 46.88$\pm$1.87&43.22$\pm$1.36\\
\bottomrule 
\end{tabular}
\end{adjustbox}
\end{table}

% However, upon thorough investigation, we have observed that (note the following experiments' observation all based on FP16 activation and INT4 weights on LLMs, see \sref{sec:experiment_setting}):
% Besides the aforementioned algorithms, there are numerous fruitful Post-Training Quantization (PTQ) works on LLMs, such as SmoothQuant~\cite{xiao2022smoothquant}, AWQ~\cite{lin2023awq}, Quip~\cite{chee2023quip}, SqueezeLLM~\cite{kim2023squeezellm}, QUIK~\cite{ashkboos2023towards}, and LLM-FP4~\cite{liu2023llm}. However, they either require extra sparsity matrices or extra steps to identify the sentitive weights. Moreover, most existing works primarily focus on zero-shot perplexity metrics and assert a strong correlation between perplexity and zero-shot accuracy performance~\cite{yao2023zeroquant,frantar2022gptq,wu2023zeroquant}. The generalizability of these conclusions to other generation tasks remains an open question.

In addition to the algorithms previously mentioned, there has been significant progress in Post-Training Quantization (PTQ) for LLMs, highlighted by innovations such as SmoothQuant~\cite{xiao2022smoothquant}, AWQ~\cite{lin2023awq}, Quip~\cite{chee2023quip}, SqueezeLLM~\cite{kim2023squeezellm}, QUIK~\cite{ashkboos2023towards}, and LLM-FP4~\cite{liu2023llm} and many more~\cite{yao2023zeroquant,dettmers2022case}. These methodologies, however, often necessitate the use of additional sparsity matrices or extra procedures to pinpoint sensitive weights. Furthermore, the majority of these studies concentrate predominantly on zero-shot perplexity and accuracy performance~\cite{yao2023zeroquant,frantar2022gptq,wu2023zeroquant}. Yet, the extent to which these findings can be generalized to other generative tasks remains to be fully explored.

\paragraph{Experiment Settings}\label{sec:experiment_setting}
 We assess performance across three metrics: Zero-Shot tasks, Code Generation, and Summarization. We also perform try to implement comparative experiments for those chat-based models and judged by GPT-4 based on the FastChat codes~\cite{zheng2023judging}. Despite this, due to significant variability in our findings, we concluded that there is no clear link between bit precision and performance. These results are detailed further in the Appendix of our study.

% However, due to significant variability in the results, we do not think their score indicate a strong relation between the bits and performance, and thus relegated to the Appendix.

\begin{itemize}
    \item \textbf{Zero-Shot Tasks.} Leveraging open-source repositories\footnote{\url{https://github.com/microsoft/DeepSpeed/tree/master/deepspeed/compression}, \url{https://github.com/qwopqwop200/GPTQ-for-LLaMa}, and \url{https://github.com/jerry-chee/QuIP}}, we applied GPTQ quantization algorithms to measure both perplexity and accuracy in zero-shot contexts. The datasets used for perplexity measurement include PTB~\cite{marcinkiewicz1994building}, Wikitext~\cite{merity2016pointer}, and C4~\cite{raffel2020exploring}.\footnote{Following the approach in \href{https://github.com/qwopqwop200/GPTQ-for-LLaMa}{gptq-for-llama}, we added two new validation sets: PTB-new, using the PTB test dataset, and C4-new, comprising the first 256$\times$seqlength. These new sets are implemented as per \href{https://github.com/jerry-chee/QuIP}{QuIP}.} For accuracy, we randomly pick ten tasks: ARC (Challenge/Easy)~\cite{boratko2018systematic}, BoolQ~\cite{clark2019boolq}, CB~\cite{de2019commitmentbank}, Copa~\cite{afshar2018copa}, PIQA~\cite{tata2003piqa}, RTE~\cite{dagan2013recognizing}, WSC~\cite{levesque2012winograd}, Storycloze~\cite{mostafazadeh2017lsdsem}). Calibration for GPTQ used 128 (32) samples for LLaMa-1B/13B (-65B) models.\footnote{\llama-13B/65B are from \cite{touvron2023llama} and \llama-1B is from \cite{xia2023sheared}. They can be downloaded from HuggingFace with names: `princeton-nlp/Sheared-LLaMA-1.3B', `huggyllama/llama-13b', `huggyllama/llama-65b'. }. We believe the results generalize to other models  family such sh OPT~\cite{zhang2022opt} and BLOOM~\cite{scao2022bloom}, The experiments were deterministic, using the seed 123.

    \item \textbf{Code Generation.} Following \cite{zheng2023codegeex} and their open-source implementation\footnote{\url{https://github.com/THUDM/CodeGeeX2}}, we adapted non-greedy generation settings (n=20, t=0.2, top\_p=0.95). To mitigate variance, nine random seeds \{111,222,\ldots, 888, 1111\} were employed. The models evaluated included CodeGeeX2-6B, \codestar-15B~\cite{li2023starcoder}, and \codellama-34B~\cite{mftcoder2023}.\footnote{Available as `THUDM/codegeex2-6b', `bigcode/starcoder', and `codefuse-ai/CodeFuse-CodeLlama-34B' on HuggingFace.} We focused on Python and JavaScript, noting instability in other programming languages.

    \item \textbf{Summarization Tasks.} Based on \cite{li2022dqbart,wu2023understanding} and their open-source codes,\footnote{\url{https://github.com/amazon-science/dq-bart}} we utilized BART-large, fine-tuned for CNNDailyMail~\cite{hermann2015teaching} and XSum~\cite{narayan2018don} summarization tasks.\footnote{Models available as `facebook/bart-large-cnn' and `facebook/bart-large-xsum' on HuggingFace.} Default settings were applied for all other parameters.
\end{itemize}

% Besides the above algorithm mentioned, there are many more fruitful PTQ works on LLMs such as SmoothQuant~\cite{xiao2022smoothquant},AWQ~\cite{lin2023awq}, Quip~\cite{chee2023quip}, SqueezeLLM~\cite{kim2023squeezellm}, QUIK~\cite{ashkboos2023towards}, LLM-FP4~\cite{liu2023llm}, etc. However, most of the existing works focus only on showing the zero-shot perplexity metrics and claim a high relationship between perplexity and zero-shot accuracy performance \cite{yao2023zeroquant,frantar2022gptq,wu2023zeroquant}. Whether the conclusions based on perplexity can generalize to other generation tasks remains unknown. 

We focus on thes experiments with FP16 activation and INT4 weights on LLMs. Our experimental setup includes a single-GPU environment, utilizing either a V100-32g or H100-80g GPU. Based on the results \tref{table:ppl-int4}, \tref{table:pp-int4a}, and \tref{table:zero-shot-int4}, we make the following two key observations. 
 \paragraph{GPTQ's Tendency to Overfit.} Although GPTQ is innovative in post-training quantization, it tends to overfit to particular datasets, especially noticeable in its fine-grain quantization results. As indicated in  \tref{table:ppl-int4}, we see that if we calibrate with specific dataset such as C4 for GPTQ, then the performance for this C4 dataset would be much better (see 9.34 or 6.74 using FGQ), while other datasets such as PTB would result in much worse performance (see 1399.89 and 22.14 using FGQ). Independently, \cite{williams2023does} also notices this issue while examining  \llama-7B.

It is admitted that the over-fitting phenomena is less severe for larger models (moving from 1B to 13B or 65B). Indeed, as shown in \tref{table:pp-int4a}, we see that \llama-65B using GPTQ on FGQ for INT4-weight results in the best average perplexity 6.61 comparing to RTN (7.17), much closer 
 to the baseline 6.41. However, its effectiveness in enhancing Zero-Shot performance is somewhat limited (detailed in  \tref{table:zero-shot-int4}), suggesting a gap in its adaptability across various language modeling scenarios and highlighting the need for robustness in model evaluation. In particularly, we presents in  \tref{table:zero-shot-int4}  the comparison between \textit{RTN} and \textit{GPTQ} on INT4 weight while keep the activation untouched, we can not claim that  \textit{GPTQ}  and \textit{RTN} are better than another based on zero-shot performance. In fact, for \llama-65B, the performance for \textit{RTN} is surprisingly better than the one of FP16.

 % \item \textbf{More tests needed for Generative Tasks:}    
    % GPTQ is overfitted to a specific dataset, particular for the fine-grain quantization (Table \ref{table:ppl-int4}). Further more, it didn't improve too much for the Zero-Shot performance (Table \ref{table:acc-int4})
 \paragraph{Expanding Evaluation Methods for Generative Models.} Our current analysis, mainly centered on zero-shot performance as shown in \tref{table:pp-int4a} and \tref{table:zero-shot-int4}, highlights the need for a broader scope in evaluation techniques. The core strength of LLMs lies in their ability to generate sequences. Therefore, this paper focuses on assessing summarization and code generation, as elaborated in \tref{table:gen-int4a}.\footnote{It should be noted that GPTQ tends to overfit to the calibrated dataset and poses implementation challenges, leading us to solely utilize RTN for our evaluations.} This strategy underlines the importance of comprehensive and detailed testing methods that extend beyond zero-shot learning, aiming to fully evaluate the generative capabilities of LLMs. The data in \tref{table:gen-int4a} show a notable difference in performance with INT4, especially when compared to standard benchmarks. For example, the performance of the \codellama-34B model in Java-Script drops from 45.05 (FP16) to 43.45 (INT4, CGQ) or 43.22 (INT4, FGQ), a decrease of 1.6 and 1.83 points, respectively. While FGQ on INT4 offers considerable improvements over CGQ, gaps compared to FP16 persist, particularly for smaller models and in Java Scripts. Interestingly, the INT4 \codellama-34B on FGQ achieves 46.88 in Python code, surpassing its baseline, whereas the INT4 \codegee-6B on FGQ scores only 29.8, falling behind even its INT4-CGQ performance. This highlights the inconsistency of INT4.
 
 These results emphasize the need for research into the effectiveness of INT4 in complex generative tasks.

% \begin{table}[H]
% \caption{\textbf{Abstractive Summarization}. Compare FP6 and INT4 weight on BART-406M for the two datasets: CNN (Left) and XSUM (Right), with per-row quantization using RTN algorithm.}
% \begin{adjustbox}{width=0.999\linewidth}
% \begin{tabular}{l|c|c}\toprule 
%      & \bart-CNN & \bart-XSUM \\
%         & Rouge-1/Rouge-2/Rouge-L/Rouge-Lsum  & Rouge-1/Rouge-2/Rouge-L/Rouge-Lsum  \\\midrule 
%  FP16       & 44.07/21.09/30.65/41.02      & 45.49/22.39/37.28/37.28      \\
% INT4     & 43.67/20.68/30.07/40.62      & 43.76/20.76/35.82/35.82  \\ 
% % 2 & \ourformat & 44.06   & 21.07   & 30.67   & 40.99      & 45.37    & 22.20   & 37.11   & 37.11      \\
% \bottomrule 
% \end{tabular}
% \end{adjustbox}
% \end{table}

% \subsection{Conclusion}
% Given Result 1/2, we know a comprehensive evaluation pipeline is needed. 
\section{Sweet Spot Solution: FP6}
\label{sec:methodology}
% Floating point quantization as a non-uniform mapping strategy of the weight or activation have recently gain attention for LLMs \cite{wu2023zeroquant, liu2023llm,zhang2023integer,micikevicius2022fp8,cambier2020shifted,kuzmin2022fp8,van2023fp8}. In fact, FP8 in activation has proven to be highly better than INT8~\cite{wu2023zeroquant}.  It's common sense that FP6 as higher bit would be better than INT4 but how much better and how much the FP6 robust to the quantilization algorithms remaisn to be measure.  

Building on previous discussions around the challenges and limitations associated with INT4 quantization, particularly its instability and subpar outcomes in code generation and summarization tasks, this section delves into an emerging area of interest in floating point quantization research. Recent studies have increasingly focused on the use of floating point quantization for handling weights or activations within LLMs \cite{wu2023zeroquant, liu2023llm,zhang2023integer,micikevicius2022fp8,cambier2020shifted,kuzmin2022fp8,van2023fp8}. Notably, a simple FP8's application  in activation processes has shown remarkable improvements over the use of INT8 \cite{wu2023zeroquant}. Inspired by these advancements, a critical question arises: \emph{Could increasing the bit precision, for instance to 5 or 6 bits, offer more stable and robust outcomes in generative tasks? This section aims to explore the extent of FP6's (FP5's) effectiveness and its resilience to different quantization algorithms, offering a potential solution to the dilemma posed by previous INT4 quantization challenges.}

For completeness, we provide a simplified overview of the floating-point format. For a detailed explanation, please refer to~\cite{wiki:floatarith}. 
A standard floating point number comprises three parts: the sign bit, the exponent bits, and the mantissa bits. This can be simplified as:
\begin{equation}
    x = S \times 2^{E-b} \times M,
\end{equation}
where \(S\) represents the sign (\(\pm 1\)), \(E\) denotes the exponent (\([0, 2^e-1]\), with \(e\) being the count of exponent bits), \(b\) is the bias for the exponent (usually \(2^{e-1}-1\)), and \(M\) signifies the mantissa (\([0, 2)\)). 

Following the implementation of \cite{zhang2019qpytorch}, the maximum/minimum achievable value in \ourformat is \(\pm28\) (\(\pm1 \times 2^{4} \times 1.75\)).\footnote{Note that for some other work such as \cite{micikevicius2022fp8} has specific configuration defined for exceptional values like NaN and \(\pm\infty\). However, these are not included in our weight quantization process using \ourformat. We do not think this slight difference will greatly impact the model performance.}  The FP16 (or BF16) weight matrix undergoes quantization as follows:
\begin{equation}
\label{eq:quant1}
    \hat W_{fp16} \approx Quant(W_{fp16}) = S_{fp16} \times W_{fp6},
\end{equation}
where \(W_{fp16}\) is the original full precision weight, \(Quant(\cdot)\) symbolizes the quantization function, \(S_{fp16}\) is the scaling factor, and \(W_{fp6}\) is the \ourformat number. 
The scaling factor \(S_{fp16}\) is computed by:
$
    S_{fp16} = {max(abs(W_{fp16}))}/{28},
$
thereby ensuring optimal use of \(W_{fp16}\)'s range without compromising on precision. Please see~\sref{sec:system} for additional customizations in \ourformat. Similar settting is defined for F5$_{E3M1}$.

\noindent\textbf{Why not INT6 instead of FP6.} The choice of FP6 over INT6 is driven by two key factors: firstly, the FP format simplifies conversion processes, as final computations are typically performed using FP16 or BF16. Secondly, there is no observed difference in accuracy between these formats, as supported by findings in \cite{wu2023zeroquant} , eliminating the need for additional experimental validation.

\subsection{Results of FP6 and FP5 on all tasks}

\begin{table}[H]
\caption{\textbf{Generation Tasks (Rouge or Pass@1 $\uparrow$)}. Comparative results of quantizations on \bart, \codegee-6B, \codestar-15B, and \codellama-34B models (left to right). Summarization tasks using two \bart versions fine-tuned by CNN/XSUM and code generation tasks in Python and JavaScript, averaged over 8 iterations with standard deviation. FP6 (FP5) format is E3M2 (E3M1).} \label{table:gen-all}
\begin{adjustbox}{width=0.999\linewidth}
\begin{tabular}{l|c|cc|cc|cc}\toprule 
             & \multicolumn{1}{c|}{\bart {\small (R1/R2/RL)}}    & \multicolumn{2}{c|}{\codegee-6B {\small (pass@1)}}               & \multicolumn{2}{c|}{\codestar-15B {\small(pass@1)}}    & \multicolumn{2}{c}{\codellama-34B {\small(pass@1)}}             \\
   Precision (RTN)   &  XSUM (CNN)  & Python        & Java-Scrpit       & Python        & Java-Scrpit  & Python        & Java-Scrpit          \\\midrule 
%  FP16      & 44.07/21.09/30.65/41.02      & 45.49/22.39/37.28/37.28 & 34.04$\pm$1.70 & 31.50$\pm$2.62  & 35.43$\pm$2.21  & 33.67$\pm$2.02 \\
% INT4       &  43.67/20.68/30.07/40.62      & 43.76/20.76/35.82/35.82 & 33.08$\pm$2.07  & 25.15$\pm$1.97 & 33.20$\pm$1.40  & 32.18$\pm$1.29  \\
% 2 & \ourformat & 34.10$\pm$2.09  & 27.74$\pm$1.18 & 31.61$\pm$1.74 & 31.15 & 35.64$\pm$1.26 & 27.91$\pm$1.49 & 33.60$\pm$1.91  & 32.38 \\

  FP16      & 45.49/22.39/37.28 (44.07/21.09/30.65) & 34.04±1.70  & 31.50±2.62  & 35.43±2.21 & 33.67±2.02 & 43.22±2.21 & 45.05±1.60  \\
% 44.06/21.07/30.669/40.9938  & 45.37/22.2/37.1095/37.1087   & 34.1±2.09  & 31.61±1.74 & 35.64±1.26 & 33.6±1.91  & 44.31±1.88 & 44.51±1.3  \\
INT4  (CGQ)     & 43.76/20.76/35.82 (43.67/20.68/30.07) & 33.08±2.07 & 25.15±1.97 & 33.20±1.40   & 32.18±1.29 & 39.84±1.64 & 43.45±2.05 \\
INT4  (FGQ)     & 44.82/21.63/36.48 (43.73/20.72/30.24)   &29.80$\pm$1.76 &28.35$\pm$2.36 & 35.64$\pm$2.52 & 32.32$\pm$2.01  & 46.88$\pm$1.87&43.22$\pm$1.36\\
FP5  (CGQ)     &  45.12/22.11/36.96 (44.13/21.19/30.68) & 30.56$\pm$2.12 & 28.43$\pm$2.39&34.30$\pm$1.29 &33.00$\pm$1.54  & 40.55$\pm$0.92 &43.29$\pm$1.93\\
\hc FP6  (CGQ) &  45.37/22.20/37.11 (44.06/21.07/30.67) &34.10±2.09&31.61±1.74& 35.64±1.26& 33.60±1.91& 44.31±1.88& 44.51±1.30\\
\bottomrule 
\end{tabular}
\end{adjustbox}
\\ 
\vspace{0.3cm}
\\
%\begin{table}[tb]
 \centering
 \caption{\textbf{Zero-Shot Evaluation (Perplexity$\downarrow$).} Compare between GPTQ$_{C4}$ and RTN quantization algorithms for INT4 weight (W4A16) on \llama of size 1B, 13B and 65B. We apply  fine-grain quantization (FGQ) in which the block-size is 256 elements per scale (\llama-1B's block-size is 128). We also report results for coarse-grain quantization (CGQ) (per row per scale). The evaluation datasets are  Wikitext2, PTB, C4, PTB-new, and C4-new.  FP6 (FP5) format is E3M2 (E3M1)} \label{table:ppl-all} %table:gen-all
 \begin{adjustbox}{width=0.999\linewidth}\label{table:acc-int4}
\begin{tabular}{lllll}\toprule
 Precision &   FGQ    & \llama-1B (4096-seq)                                          & \llama-13B (2048-seq)                                     & \llama-65B (2048-seq)                                  \\\midrule
FP16    &  N/A & 24.31$_{7.53/37.39/8.91/58.34/9.40 }$      & 13.16$_{5.09/19.23/6.61/28.10/6.80 }$ & 6.41$_{3.56/8.00/5.62/8.88/5.98 }$  \\\midrule
INT4-GPTQ$_{C4}$  &  \cmark &564.73$_{7.83/1399.89/9.34/1396.76/9.84 }$ & 14.19$_{5.28/22.14/6.74/29.83/6.95}$  & 6.61$_{3.81/8.17/5.73/9.20/6.13 }$  \\
INT4-RTN   &  \cmark &22.76$_{7.98/34.03/9.50/52.28/10.00 }$     & 14.32$_{5.35/22.49/6.80/29.96/7.00 }$ & 7.30$_{3.79/10.13/5.74/10.54/6.31}$  \\
 INT4-GPTQ$_{C4}$  &  \xmark &288.22$_{8.20/719.21/9.84/693.48/10.37 }$  & 14.13$_{5.37/21.31/6.84/30.01/7.09 }$ & 7.17$_{4.12/10.50/5.83/9.16/6.21 }$ \\
 INT4-RTN   & \xmark & 34.29$_{8.33/55.52/10.15/86.85/10.62 }$    & 14.32$_{5.55/20.95/6.97/30.91/7.22 }$ & 7.72$_{4.20/10.59/5.90/11.44/6.47}$\\\midrule
  FP5-GPTQ$_{C4}$  &  \cmark & 44.29$_{7.74/72.63/9.19/122.20/9.68 }$ & 13.76$_{5.22/20.43/6.71/29.53/6.92 }$ &  6.50$_{3.67/8.15/5.68/8.95/6.04 }$\\
   FP5-RTN   &  \cmark &28.52$_{7.78/44.09/9.23/71.79/9.71 }$ & 13.95$_{5.20/21.01/6.71/29.92/6.92 }$&  6.83$_{3.70/9.73/5.69/8.98/6.06 }$\\
 \gb FP5-GPTQ$_{C4}$  &  \xmark &32.03$_{7.77/50.66/9.23/82.73/9.76 }$ &13.90$_{5.22/21.02/6.71/29.64/6.93 }$ &  6.46$_{3.68/7.83/5.70/9.13/5.98 }$\\
\gc FP5-RTN   &  \xmark  &27.92$_{7.83/41.80/9.27/70.94/9.77 }$ & 14.10$_{5.22/21.52/6.72/30.13/6.93 }$  &  6.50$_{3.68/8.08/5.69/8.98/6.06 }$\\  \midrule
FP6-GPTQ$_{C4}$  &  \cmark &22.77$_{7.59/34.04/8.98/53.76/9.47 }$ & 13.36$_{5.13/19.67/6.63/28.53/6.83 }$ & 6.47$_{3.59/8.12/5.63/8.91/6.10 }$ \\
FP6-RTN   &  \cmark &23.42$_{7.60/35.84/8.99/55.20/9.49 }$                                & 13.24$_{5.12/19.43/6.63/28.19/6.83 }$                                & 6.44$_{3.58/8.04/5.63/8.92/6.05 }$ \\  
\hb FP6-GPTQ$_{C4}$  &  \xmark &23.58$_{7.59/35.76/8.98/56.10/9.47 }$ & 13.23$_{5.12/19.34/6.64/28.20/6.83 }$ & 6.42$_{3.61/8.01/5.63/8.87/6.00 }$ \\
\hc FP6-RTN   &  \xmark &24.83$_{7.60/38.79/8.99/59.26/9.49 }$ & 13.09$_{5.12/19.06/6.64/27.81/6.83 }$ & 6.42$_{3.59/8.01/5.63/8.89/5.99 }$ \\  

\bottomrule  
\end{tabular}
\end{adjustbox}
 \end{table}

We conduct the experiments described in \sref{sec:experiment_setting}, namely code generation, summarization and zero-shot experiments. The results are shown in \tref{table:gen-all} and \tref{table:ppl-all}. Additional results are presented in \tref{table:fastchat} and \tref{table:acc-int-all-app}, which we defer to appendix as it we do not find a clear link between bit precision and performance.

In general, the FP6 quantization method, particularly with CGQ, stands out in this analysis, offering a balance of high performance and robustness across different tasks, models and even quantiztion algorithms (RTN and GPTQ), a notable improvement over both FP5 and INT4 quantizations. In addition, we make the following observation:

%\label{table:ppl-all} %table:gen-all

% \textbf{FP6 on CGQ Performance Comparable to Baseline.} As indicated in Tables \ref{table:gen-all} and \ref{table:ppl-all}, FP6 employing CGQ (highlighted in dark orange) successfully attains accuracy levels equivalent to those of full precision.  Additionally, FP5 results were evaluated using CGQ, revealing a disparity compared to the baseline. 
 
% In the comparative analysis focusing on FP5, FP6, and INT4 quantization methods, distinct differences and outcomes were observed:

\textbf{FP5 Performance.} FP5 with CGQ shows an improvement over INT4 quantization but still does not reach the high performance levels of FP16. The gap between FP5 and its baseline is particularly noticeable in the Python and JavaScript code generation tasks across \codegee-6B, \codestar-15B, and \codellama-34B.

\textbf{FP6 Robustness.} FP6 quantization, especially with CGQ, demonstrates a significant advancement, nearly matching the FP16 baseline across various tasks and models. This quantization method not only narrows the performance gap seen in FP5 and INT4 but also shows robustness in handling different tasks. The robustness is further accentuated when comparing CGQ and FGQ within the FP6 framework (as there is little difference between CGQ and FGQ), where FP6 with CGQ consistently maintains high performance close to baseline, indicating its effectiveness and stability across different scenarios.  Moreover, FP6 is robust to quantization algorithms: either RTN or GPTQ results in similar results, particularly for \codellama-34B, as shwon in \tref{table:ppl-all}.

% INT4 Quantization Variants: Comparing INT4 variants, both CGQ and Fine Grain Quantization (FGQ) show a drop in performance when compared to FP5 and FP6. The performance decrease is more pronounced with FGQ, especially in code generation tasks in Python and JavaScript.

% \paragraph{Comparative Analysis Across Tasks.} The analysis reveals that FP6 with CGQ not only performs well in comparison to INT4 and FP5 but also demonstrates a level of robustness across various tasks, which is not as prominent in FP5 and INT4, especially with FGQ.

\section{System Support Discussion}
\label{sec:system}

\subsection{4+2 format for FP6}
\label{sec:4+2}
% Implementing and fully utilizing a 6-bit number format presents unique challenges since it does not conform to the standard power-of-2 numerical format, which we refer to as an "odd bit precision setting." There are two primary methods to address this. (1) One straightforward approach is to convert the 6-bit format into an 8-bit floating-point (FP8) format. However, this method negates the primary advantage of the 6-bit format: its ability to save memory. (2) An alternative method involves grouping multiple 6-bit numbers (e.g., 16 of them) in a contiguous memory space. These groups can then be represented using three 32-bit integers (INT32) or a 32-bit floating-point (FP32) format. This method retains the memory-saving benefit. The downside is that during the dequantization process, the task of converting these 6-bit values, which are spread across two INT32s, is significantly complex and resource-intensive.

In addressing the challenges of utilizing a non-standard 6-bit number format, which deviates from the conventional power-of-2 numerical systems (termed as "odd bit precision setting"), we propose a novel approach. This method is distinct from the two commonly considered strategies:

\begin{enumerate}
  \item The first approach involves directly converting the 6-bit format into an 8-bit floating-point (FP8) format. While this is a straightforward solution, it unfortunately negates the primary benefit of the 6-bit format, which is to conserve memory.
\item The second approach entails grouping several 6-bit numbers together in a continuous memory block and representing them using either 32-bit integers (INT32) or a 32-bit floating-point (FP32) format. This method maintains the memory-saving advantage but adds complexity to the dequantization process.
\end{enumerate}

Our unique strategy, however, focuses on dividing the 6-bit number into two distinct sub-numbers: the first sub-number representing the initial 4 bits, and the second sub-number accounting for the remaining 2 bits. Our poposed "4+2" method can be seen as an advanced variation of the second standard approach.
The 4+2 bit division is based on the fundamental principle that any positive integer can be expressed as a sum of powers of 2. With this foundation, we divide the 6-bit number into two components:
\begin{itemize}
    \item The first part, comprising the initial 4 bits, handles elements such as the sign bit and a 3-bit exponent.
    \item The second part, containing the remaining 2 bits, is dedicated to the 2-bit mantissa.
\end{itemize}

This division into 4+2 bits facilitates simultaneous loading and dequantization of these sub-numbers, culminating in the generation of the final 16-bit floating-point (FP16) weight. Our approach innovatively balances the need for reduced memory footprint with the practicalities of dequantization, particularly in addressing the challenges of memory access across segmented numbers.

% To materialize and maximize the benefits of  6-bit number is challenging as it is  not a number under the power of 2 format (we call this an odd bit precision setting). 
% There are two types of way to represent it. 
% A simple and straight-forward way is to cast the \ourformat to FP8 format. 
% However, this fully eliminates the benefits of the extra memory saving of \ourformat. 
% Another way to realize this is to concatenate multiple FP6 (e.g., 16) number in a continuous memory space, and then use the corresponding INT32 (or FP32) format (3 INT32 numbers) to represent them. 
% The benefit of the second approach is that it fully achieves the memory footprint saving. 
% However, the drawback is that during the dequantization step, the overhead of dequantizing the FP6 number across two INT32 numbers is non-trivial. 

% Thanks to the convinient conversation between decimal system and binary system, we know any positive integer number can be represented by a sum of a sequence of number in the power of 2 format, e.g., $5=1+4$ and $6=4+2$. 
% Therefore, we can represent our FP6 number in 2 sub-numbers, one to store the first 4 bits of the FP6 number (e.g., the sign and the 3-bit exponent parts), and the other to store the final 2 bits of the FP6 number (e.g., the 2-bit mantissa). 

% On the fly, we load the two sub-numbers simultaneously and dequantize them together to get the final FP16 weight.
% As such, we realize the memory footprint saving and resolve the across-number memory access during dequantization. 

\subsection{Bias Shift}
\label{sec:BiasShift}

Dequantizing FP6 to FP16 during runtime on GPUs can be significantly resource-intensive, primarily due to the complexity involved in manipulating the exponent field, as detailed in Section~\ref{sec:methodology}. 

The bias term for the exponent, typically determined by the exponent bits, is 3 for FP6 and 15 for FP16. Mathematically, the process of dequantizing FP6 to FP16 (excluding the sign) is represented as:
\begin{equation}
     2^{E_{\text{FP16}} - 15} \times M_{\text{FP16}} =  2^{E_{\text{FP6}} - 3} \times M_{\text{FP6}},
\end{equation}
where the superscripts \texttt{FP16/FP6} indicate the respective format. 
It is noteworthy that  the scaling factor dequantization can be done after matrix multiplication before the accumulation for fine-grained (sub-row) quantization schemes or after accumulation for coarse-grained (row-wise) quantization schemes. 

% It is noteworthy that the scaling factor \(S_{\text{FP16}}\) from \eref{eq:quant} can be applied post-matrix multiplication (for fine-grained quantization scheme) or post-accumulation (for coarse-grained scheme) in the dequantization process. 

While padding can easily adjust the mantissa, aligning the exponents requires more effort due to the difference in biases. An analogy can be drawn with converting an INT4 number back to a symmetric INT8 format: if INT4 employs a symmetric format (for the mantissa), zero padding suffices. However, in an asymmetric format, padding alone is inadequate, and additional steps are necessary. 

To address this, we have customized our \texttt{FP6} format with a non-standard exponent bias of 15. This modification does not affect precision or accuracy because:
\begin{equation}
    S_{\text{FP16}} \times 2^{E-3} \times M = (S_{\text{FP16}} \times 2^{12})  \times 2^{E-15} \times M,
\end{equation}
meaning the bias shift can be seamlessly integrated into the scaling factor. Crucially, since \(S_{\text{FP16}}\) is less than 1, multiplying it with \(2^{12}\) still allows for accurate representation in FP16 format through simple exponent bit shifting, avoiding numerical errors.
\begin{figure}[H]
    \centering
    \subfloat[Before Bias Shift.\label{fig:BeforeShift}]{
        \includegraphics[width=0.4\linewidth]{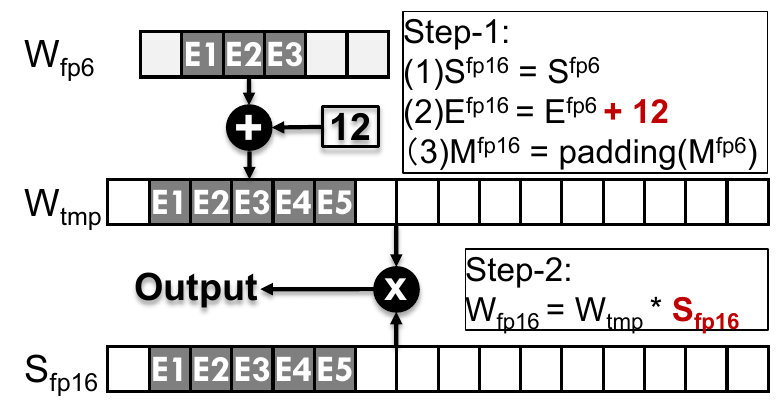}
    }
    \hspace{0.4cm}
    \subfloat[After Bias Shift.\label{fig:AfterShift}]{
        \includegraphics[width=0.4\linewidth]{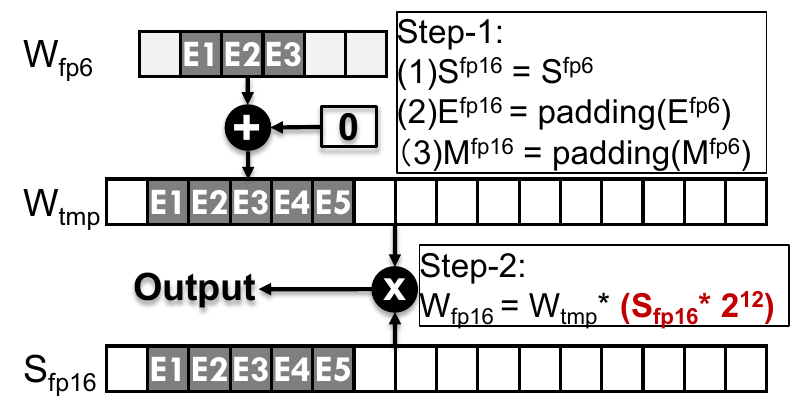}
    }
    \caption{Comparison of the Traditional Method (Left) versus Our Proposed Method (Right): Demonstrating the Significant Run-time Advantages of Bias Shift.}
    \label{fig:RuntimeBenefits}
\end{figure}
Our bias shift method greatly simplifies the FP6-FP16 de-quantization process during runtime. To demonstrate this, we provide a side-by-side comparison in Figures~\ref{fig:BeforeShift} and \ref{fig:AfterShift}.

Figure~\ref{fig:BeforeShift} outlines the original two-step process for de-quantizing each FP6 weight. The first step involves casting \( W_{\text{fp6}} \) to \( W_{\text{tmp}} \), and the second step requires multiplying by the quantization scale \( S_{\text{fp16}} \). The most demanding part of this procedure is recalculating the exponent for \( W_{\text{tmp}} \), which involves extracting \( E_{\text{fp6}} \) from \( W_{\text{fp6}} \), adding 12, and then incorporating this back into \( W_{\text{tmp}} \). Additionally, the process to de-quantize subnormal FP6 numbers\footnote{Subnormal numbers in floating-point formats are very small values, including zero, characterized by an exponent field of all zeros. When de-quantizing subnormal FP6 numbers, the new exponent should remain at zero, not adding 12, to maintain its subnormal status. Subsequently, FP16 is multiplied by \(1.0 \times 2^{12}\).} adds further complexity to the de-quantization during runtime.

However, with our bias shift strategy, as illustrated in Figure~\ref{fig:AfterShift}, the exponent adjustment becomes a straightforward bit-level padding process. The addition of the constant integer 12, initially required in Step 1, can now be deferred to Step 2, \textbf{eliminating any runtime overhead}. This is possible because the multiplication of the quantization scales with the constant integer can be performed statically after the model is quantized and before runtime. Moreover, this streamlined approach also efficiently accommodates the de-quantization of subnormal numbers.

\subsection{System evaluation}
%\zhewei{@haojun, I prefer batch size 1 setting if it does not work. Please use batch size 8. Let's just show GeMM/MatMul performance (Pure FP16 vs. FP6)for n by n matrix from 1024 to 16384. Let's say this is just a illustration, the full system implemenation is left for future work. After the other work got published, cite it here for Version-II and say we found a work fully optimized the idea and extends it in a comprehensive way.}
We conducted an evaluation of the latest GPU kernels across various weight-only quantization techniques, focusing on their system performance. This kernel-level assessment was carried out on the NVIDIA A100-40GB platform, running Linux 5.3.18 and CUDA 11.8. Our primary focus was on the performance of feed-forward (FFN) layers within the \llama models, specifically during the token generation phase, as detailed in \cite{touvron2023llama}. Comprehensive data on matrix shapes and kernel latency is available in Appendix \ref{sec:DetailedKernelLatency}. We employed cuBLAS~\cite{cuBLAS} as our benchmark for non-quantized performance (W16A16). We also included cutting-edge kernel support for F INT4 FGQ quantization (W4A16) from TensorRT-LLM \cite{TensorRT-LLM} for comparative analysis.\footnote{Between the supported block-size: 64 and 128, we chose 128.}

\begin{figure}[H]
    \centering
    \vspace{0.5cm}
    \includegraphics[width=0.7\linewidth]{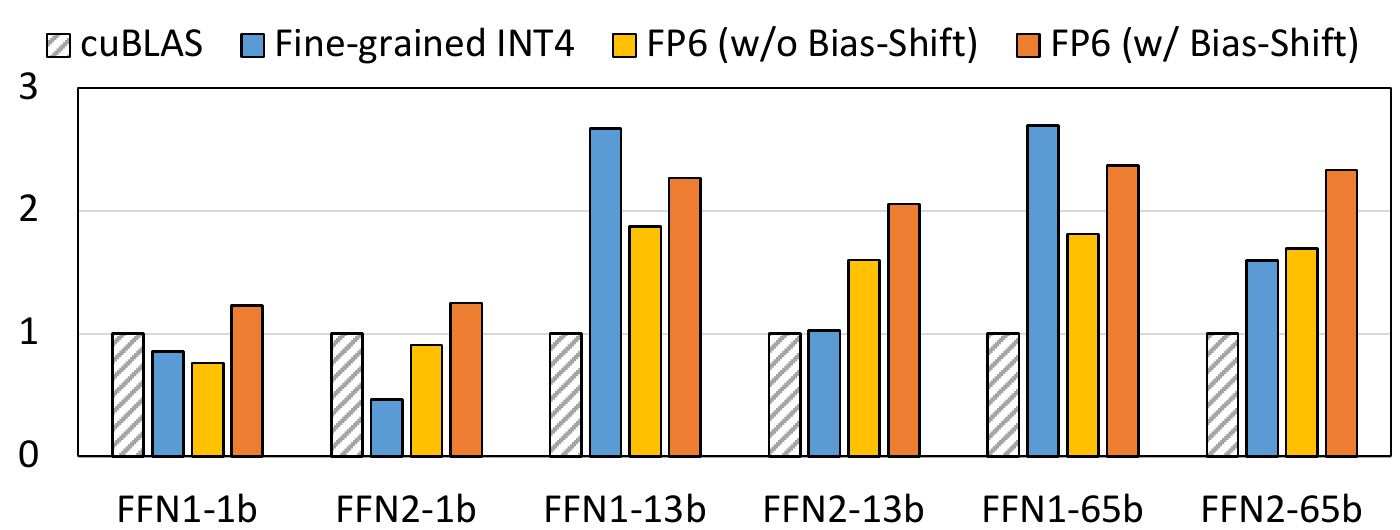}
    \caption{GPU Kernel Speedups compared to cuBLAS. FFN1 and FFN2 are defined for matrices of $4H\times H$ and $H \times 4H$ size, where $H$ are defined by the size of the \llama models.  \llama-1B, 13B and 65B are respectively $5504$, $13824$ and $22016$. See detailed number in \tref{table:kernel} }
    \label{fig:KernelPerformance}
\end{figure}
To elucidate the advantages of our proposed \textit{Bias Shift} technique, detailed in Section \ref{sec:BiasShift}, we also developed and tested an FP6 GPU kernel without \textit{Bias Shift}.

According to the results presented in Figure \ref{fig:KernelPerformance}, our FP6 kernel, enhanced with Bias-Shift, achieved speeds up to $2.37\times$ faster and on average $1.92\times$ faster than cuBLAS. Given that LLM inference is commonly constrained by limited GPU DRAM \cite{xia2023flash, kim2023full}, our approach effectively alleviates this bottleneck by minimizing model weight memory access. Moreover, our FP6 kernel outperforms the state-of-the-art fine-grained INT4 implementation in speed, being $1.06\times$ faster on average for the first feed-forward layers (FFN1) and $2.05\times$ faster for the second feed-forward layers (FFN2). It should be noted that Figure \ref{fig:KernelPerformance} provides only a snapshot of key outcomes, with a comprehensive system implementation and detailed performance analysis reserved for future work. Importantly, our FP6 kernel with Bias-Shift is on average $1.36\times$ faster compared to the same FP6 kernel without \textit{Bias Shift}, underscoring the crucial role of \textit{Bias-Shift} as discussed in Section \ref{sec:BiasShift}.

\section{Discussion and Conclusion}
\label{sec:discussion}
% Mainly diccuss the shortcoming and big-scope of the paper: 
% (1) the evaluation is still not comprehensive. A lot of generation tasks and accuracy-based tasks are not done.

% (2) the system evaluation is naive, did not compare with sota framework.

% (3) Our section-4 technique as mentioned in zeroquant-v2 also works for other bits for both bit-precision and bias-shift. Therefore, if one day, 5-bits works, out method can be simply transfered to 5-bits as well.

% (4) look forward to more advanced PTQ methods and its combination with other techniques. pace the road for the 3D decomposition project.
% \section{Discussion}

Our paper introduces a novel approach to GPU kernel optimization, specifically targeting weight-only quantization methods. Despite the significance of our findings, they pave the way for further research and development in several areas.

\textbf{Evaluation Scope.} A primary limitation of our study is still the narrow scope of evaluation although we have span to code generation and summarization. We suggest a vast field for future research. Expanding the scope to include diverse tasks and a focus on both performance and accuracy could enhance the robustness of our methods.

\textbf{Comparative Analysis.} While our system evaluation provides valuable insights, it lacks in-depth comparison with state-of-the-art (SOTA) frameworks. A more comprehensive benchmarking against advanced frameworks would offer a clearer perspective on our approach's efficacy and areas for improvement.

\textbf{Technique Adaptability.} A notable aspect of our work is the adaptability of our techniques, particularly in bit-precision and bias-shift. The potential to adapt our methods to emerging standards, such as 5-bit quantization, demonstrates their flexibility and future applicability in various contexts.

\textbf{Future Directions.} The advancement of Post-Training Quantization (PTQ) methods and their integration with other techniques presents exciting future possibilities. Our research lays the foundation for further advancements in model optimization, such as the 3-bit precision from evolving quantization techniques.

In conclusion, our study marks a significant contribution to odd-bit GPU kernel optimization. It also opens avenues for broader research, exploring the full potential of model optimization and quantization across diverse applications.

\section*{Contributions}
% Xiaoxia Wu developed the measurement of model quality and led the full paper writing. Haojun Xia, contributing to the system section writing, is the main designer for developing FP6 acceleration strategy and sophisticated FP6 system support. Stephen Youn, Zhen Zheng and Shiyang Chen   are responsible for DeepSpeed integration. Mike and Arash provide framework testing and external integration. Yuxiong He and Olatunji Ruwase provide valuable discussion. 
% Leon Song is the overall leader for algorithm and system design.  Zhewei Yao initiated this project and was instrumental in the design of the FP6/5 format.

Xiaoxia Wu led model quality measurement and paper writing. Haojun Xia, contributing to system section writing, designed the FP6 acceleration strategy and system support. Stephen Youn, Zhen Zheng, and Shiyang Chen handled DeepSpeed integration.  Arash Bakhtiari and Michael Wyatt managed framework testing and external integration. Reza Yazdani Aminabadi, Yuxiong He and Olatunji Ruwase provided key discussions. Leon Song is the overall leader for algorithm and system design. Zhewei Yao initiated the project and led the FP6/5 format design.

% \clearpage
{
\bibliographystyle{plain}
\bibliography{ref.bib}

\begin{thebibliography}{10}

\bibitem{afshar2018copa}
Ardavan Afshar, Ioakeim Perros, Evangelos~E Papalexakis, Elizabeth Searles, Joyce Ho, and Jimeng Sun.
\newblock Copa: Constrained parafac2 for sparse \& large datasets.
\newblock In {\em Proceedings of the 27th ACM International Conference on Information and Knowledge Management}, pages 793--802, 2018.

\bibitem{ashkboos2023towards}
Saleh Ashkboos, Ilia Markov, Elias Frantar, Tingxuan Zhong, Xincheng Wang, Jie Ren, Torsten Hoefler, and Dan Alistarh.
\newblock Towards end-to-end 4-bit inference on generative large language models.
\newblock {\em arXiv preprint arXiv:2310.09259}, 2023.

\bibitem{bai2020binarybert}
Haoli Bai, Wei Zhang, Lu~Hou, Lifeng Shang, Jing Jin, Xin Jiang, Qun Liu, Michael Lyu, and Irwin King.
\newblock Binarybert: Pushing the limit of bert quantization.
\newblock {\em arXiv preprint arXiv:2012.15701}, 2020.

\bibitem{boratko2018systematic}
Michael Boratko, Harshit Padigela, Divyendra Mikkilineni, Pritish Yuvraj, Rajarshi Das, Andrew McCallum, Maria Chang, Achille Fokoue-Nkoutche, Pavan Kapanipathi, Nicholas Mattei, et~al.
\newblock A systematic classification of knowledge, reasoning, and context within the arc dataset.
\newblock {\em arXiv preprint arXiv:1806.00358}, 2018.

\bibitem{brown2020language}
Tom~B Brown, Benjamin Mann, Nick Ryder, Melanie Subbiah, Jared Kaplan, Prafulla Dhariwal, Arvind Neelakantan, Pranav Shyam, Girish Sastry, Amanda Askell, et~al.
\newblock Language models are few-shot learners.
\newblock {\em arXiv preprint arXiv:2005.14165}, 2020.

\bibitem{cai2020zeroq}
Yaohui Cai, Zhewei Yao, Zhen Dong, Amir Gholami, Michael~W Mahoney, and Kurt Keutzer.
\newblock Zeroq: A novel zero shot quantization framework.
\newblock In {\em Proceedings of the IEEE/CVF Conference on Computer Vision and Pattern Recognition}, pages 13169--13178, 2020.

\bibitem{cambier2020shifted}
Léopold Cambier, Anahita Bhiwandiwalla, Ting Gong, Mehran Nekuii, Oguz~H Elibol, and Hanlin Tang.
\newblock Shifted and squeezed 8-bit floating point format for low-precision training of deep neural networks.
\newblock {\em arXiv preprint arXiv:2001.05674}, 2020.

\bibitem{chee2023quip}
Jerry Chee, Yaohui Cai, Volodymyr Kuleshov, and Christopher De~Sa.
\newblock Quip: 2-bit quantization of large language models with guarantees.
\newblock {\em arXiv preprint arXiv:2307.13304}, 2023.

\bibitem{clark2019boolq}
Christopher Clark, Kenton Lee, Ming-Wei Chang, Tom Kwiatkowski, Michael Collins, and Kristina Toutanova.
\newblock Boolq: Exploring the surprising difficulty of natural yes/no questions.
\newblock {\em arXiv preprint arXiv:1905.10044}, 2019.

\bibitem{wiki:floatarith}
Wikipedia contributors.
\newblock Floating-point arithmetic --- {Wikipedia}{,} the free encyclopedia, last edited 2023.
\newblock [Online; accessed 8-December-2023].

\bibitem{dagan2013recognizing}
Ido Dagan, Dan Roth, Mark Sammons, and Fabio~Massimo Zanzotto.
\newblock Recognizing textual entailment: Models and applications.
\newblock {\em Synthesis Lectures on Human Language Technologies}, 6(4):1--220, 2013.

\bibitem{de2019commitmentbank}
Marie-Catherine De~Marneffe, Mandy Simons, and Judith Tonhauser.
\newblock The commitmentbank: Investigating projection in naturally occurring discourse.
\newblock In {\em proceedings of Sinn und Bedeutung}, volume~23, pages 107--124, 2019.

\bibitem{dettmers2023qlora}
Tim Dettmers, Artidoro Pagnoni, Ari Holtzman, and Luke Zettlemoyer.
\newblock Qlora: Efficient finetuning of quantized llms.
\newblock {\em arXiv preprint arXiv:2305.14314}, 2023.

\bibitem{dettmers2022case}
Tim Dettmers and Luke Zettlemoyer.
\newblock The case for 4-bit precision: k-bit inference scaling laws.
\newblock {\em arXiv preprint arXiv:2212.09720}, 2022.

\bibitem{dong2019hawq}
Zhen Dong, Zhewei Yao, Amir Gholami, Michael~W Mahoney, and Kurt Keutzer.
\newblock {HAWQ}: Hessian aware quantization of neural networks with mixed-precision.
\newblock In {\em Proceedings of the IEEE International Conference on Computer Vision}, pages 293--302, 2019.

\bibitem{esser2019learned}
Steven~K Esser, Jeffrey~L McKinstry, Deepika Bablani, Rathinakumar Appuswamy, and Dharmendra~S Modha.
\newblock Learned step size quantization.
\newblock {\em arXiv preprint arXiv:1902.08153}, 2019.

\bibitem{fan2020training}
Angela Fan, Pierre Stock, Benjamin Graham, Edouard Grave, Remi Gribonval, Herve Jegou, and Armand Joulin.
\newblock Training with quantization noise for extreme fixed-point compression.
\newblock {\em arXiv preprint arXiv:2004.07320}, 2020.

\bibitem{frantar2022optimal}
Elias Frantar and Dan Alistarh.
\newblock Optimal brain compression: A framework for accurate post-training quantization and pruning.
\newblock {\em arXiv preprint arXiv:2208.11580}, 2022.

\bibitem{frantar2022gptq}
Elias Frantar, Saleh Ashkboos, Torsten Hoefler, and Dan Alistarh.
\newblock Gptq: Accurate post-training quantization for generative pre-trained transformers.
\newblock {\em arXiv preprint arXiv:2210.17323}, 2022.

\bibitem{gholami2021survey}
Amir Gholami, Sehoon Kim, Zhen Dong, Zhewei Yao, Michael~W Mahoney, and Kurt Keutzer.
\newblock A survey of quantization methods for efficient neural network inference.
\newblock {\em arXiv preprint arXiv:2103.13630}, 2021.

\bibitem{copilot}
GitHub.
\newblock Github copilot.
\newblock \url{https://github.com/features/copilot/}, 2021.

\bibitem{guo2023olive}
Cong Guo, Jiaming Tang, Weiming Hu, Jingwen Leng, Chen Zhang, Fan Yang, Yunxin Liu, Minyi Guo, and Yuhao Zhu.
\newblock Olive: Accelerating large language models via hardware-friendly outlier-victim pair quantization.
\newblock In {\em Proceedings of the 50th Annual International Symposium on Computer Architecture}, pages 1--15, 2023.

\bibitem{guo2023lq}
Han Guo, Philip Greengard, Eric~P Xing, and Yoon Kim.
\newblock Lq-lora: Low-rank plus quantized matrix decomposition for efficient language model finetuning.
\newblock {\em arXiv preprint arXiv:2311.12023}, 2023.

\bibitem{hassibi1993second}
Babak Hassibi and David~G Stork.
\newblock Second order derivatives for network pruning: Optimal brain surgeon.
\newblock In {\em Advances in neural information processing systems}, pages 164--171, 1993.

\bibitem{hermann2015teaching}
Karl~Moritz Hermann, Tomas Kocisky, Edward Grefenstette, Lasse Espeholt, Will Kay, Mustafa Suleyman, and Phil Blunsom.
\newblock Teaching machines to read and comprehend.
\newblock {\em arXiv preprint arXiv:1506.03340}, 2015.

\bibitem{hubara2017quantized}
Itay Hubara, Matthieu Courbariaux, Daniel Soudry, Ran El-Yaniv, and Yoshua Bengio.
\newblock Quantized neural networks: Training neural networks with low precision weights and activations.
\newblock {\em The Journal of Machine Learning Research}, 18(1):6869--6898, 2017.

\bibitem{kim2023memory}
Jeonghoon Kim, Jung~Hyun Lee, Sungdong Kim, Joonsuk Park, Kang~Min Yoo, Se~Jung Kwon, and Dongsoo Lee.
\newblock Memory-efficient fine-tuning of compressed large language models via sub-4-bit integer quantization.
\newblock {\em arXiv preprint arXiv:2305.14152}, 2023.

\bibitem{kim2021bert}
Sehoon Kim, Amir Gholami, Zhewei Yao, Michael~W Mahoney, and Kurt Keutzer.
\newblock I-bert: Integer-only bert quantization.
\newblock In {\em International conference on machine learning}, pages 5506--5518. PMLR, 2021.

\bibitem{kim2023squeezellm}
Sehoon Kim, Coleman Hooper, Amir Gholami, Zhen Dong, Xiuyu Li, Sheng Shen, Michael~W Mahoney, and Kurt Keutzer.
\newblock Squeezellm: Dense-and-sparse quantization.
\newblock {\em arXiv preprint arXiv:2306.07629}, 2023.

\bibitem{kim2023full}
Sehoon Kim, Coleman Hooper, Thanakul Wattanawong, Minwoo Kang, Ruohan Yan, Hasan Genc, Grace Dinh, Qijing Huang, Kurt Keutzer, Michael~W Mahoney, et~al.
\newblock Full stack optimization of transformer inference: a survey.
\newblock {\em arXiv preprint arXiv:2302.14017}, 2023.

\bibitem{krishnamoorthi2018quantizing}
Raghuraman Krishnamoorthi.
\newblock Quantizing deep convolutional networks for efficient inference: A whitepaper.
\newblock {\em arXiv preprint arXiv:1806.08342}, 2018.

\bibitem{kuzmin2022fp8}
Andrey Kuzmin, Mart Van~Baalen, Yuwei Ren, Markus Nagel, Jorn Peters, and Tijmen Blankevoort.
\newblock Fp8 quantization: The power of the exponent.
\newblock {\em arXiv preprint arXiv:2208.09225}, 2022.

\bibitem{lecun1990optimal}
Yann LeCun, John~S Denker, and Sara~A Solla.
\newblock Optimal brain damage.
\newblock In {\em Advances in neural information processing systems}, pages 598--605, 1990.

\bibitem{lee2023owq}
Changhun Lee, Jungyu Jin, Taesu Kim, Hyungjun Kim, and Eunhyeok Park.
\newblock Owq: Lessons learned from activation outliers for weight quantization in large language models.
\newblock {\em arXiv preprint arXiv:2306.02272}, 2023.

\bibitem{levesque2012winograd}
Hector Levesque, Ernest Davis, and Leora Morgenstern.
\newblock The winograd schema challenge.
\newblock In {\em Thirteenth International Conference on the Principles of Knowledge Representation and Reasoning}. Citeseer, 2012.

\bibitem{li2023starcoder}
Raymond Li, Loubna~Ben Allal, Yangtian Zi, Niklas Muennighoff, Denis Kocetkov, Chenghao Mou, Marc Marone, Christopher Akiki, Jia Li, Jenny Chim, Qian Liu, Evgenii Zheltonozhskii, Terry~Yue Zhuo, Thomas Wang, Olivier Dehaene, Mishig Davaadorj, Joel Lamy-Poirier, João Monteiro, Oleh Shliazhko, Nicolas Gontier, Nicholas Meade, Armel Zebaze, Ming-Ho Yee, Logesh~Kumar Umapathi, Jian Zhu, Benjamin Lipkin, Muhtasham Oblokulov, Zhiruo Wang, Rudra Murthy, Jason Stillerman, Siva~Sankalp Patel, Dmitry Abulkhanov, Marco Zocca, Manan Dey, Zhihan Zhang, Nour Fahmy, Urvashi Bhattacharyya, Wenhao Yu, Swayam Singh, Sasha Luccioni, Paulo Villegas, Maxim Kunakov, Fedor Zhdanov, Manuel Romero, Tony Lee, Nadav Timor, Jennifer Ding, Claire Schlesinger, Hailey Schoelkopf, Jan Ebert, Tri Dao, Mayank Mishra, Alex Gu, Jennifer Robinson, Carolyn~Jane Anderson, Brendan Dolan-Gavitt, Danish Contractor, Siva Reddy, Daniel Fried, Dzmitry Bahdanau, Yacine Jernite, Carlos~Muñoz Ferrandis, Sean Hughes, Thomas Wolf, Arjun Guha, Leandro von
  Werra, and Harm de~Vries.
\newblock Starcoder: may the source be with you!
\newblock {\em 2305.06161}, 2023.

\bibitem{li2022dqbart}
Zheng Li, Zijian Wang, Ming Tan, Ramesh Nallapati, Parminder Bhatia, Andrew Arnold, Bing Xiang, and Dan Roth.
\newblock Dq-bart: Efficient sequence-to-sequence model via joint distillation and quantization.
\newblock In {\em Proceedings of the 60th Annual Meeting of the Association for Computational Linguistics (Volume 2: Short Papers)}, pages 203--211, 2022.

\bibitem{lin2023awq}
Ji~Lin, Jiaming Tang, Haotian Tang, Shang Yang, Xingyu Dang, and Song Han.
\newblock Awq: Activation-aware weight quantization for llm compression and acceleration.
\newblock {\em arXiv preprint arXiv:2306.00978}, 2023.

\bibitem{mftcoder2023}
Bingchang Liu, Chaoyu Chen, Cong Liao, Zi~Gong, Huan Wang, Zhichao Lei, Ming Liang, Dajun Chen, Min Shen, Hailian Zhou, Hang Yu, and Jianguo Li.
\newblock Mftcoder: Boosting code llms with multitask fine-tuning.
\newblock {\em arXiv preprint arXiv}, 2023.

\bibitem{liu2023llm}
Shih-yang Liu, Zechun Liu, Xijie Huang, Pingcheng Dong, and Kwang-Ting Cheng.
\newblock Llm-fp4: 4-bit floating-point quantized transformers.
\newblock {\em arXiv preprint arXiv:2310.16836}, 2023.

\bibitem{liu2021post}
Zhenhua Liu, Yunhe Wang, Kai Han, Wei Zhang, Siwei Ma, and Wen Gao.
\newblock Post-training quantization for vision transformer.
\newblock {\em Advances in Neural Information Processing Systems}, 34, 2021.

\bibitem{marcinkiewicz1994building}
Mary~Ann Marcinkiewicz.
\newblock Building a large annotated corpus of english: The penn treebank.
\newblock {\em Using Large Corpora}, page 273, 1994.

\bibitem{merity2016pointer}
Stephen Merity, Caiming Xiong, James Bradbury, and Richard Socher.
\newblock Pointer sentinel mixture models.
\newblock In {\em International Conference on Learning Representations}, 2017.

\bibitem{micikevicius2022fp8}
Paulius Micikevicius, Dusan Stosic, Neil Burgess, Marius Cornea, Pradeep Dubey, Richard Grisenthwaite, Sangwon Ha, Alexander Heinecke, Patrick Judd, John Kamalu, et~al.
\newblock Fp8 formats for deep learning.
\newblock {\em arXiv preprint arXiv:2209.05433}, 2022.

\bibitem{mostafazadeh2017lsdsem}
Nasrin Mostafazadeh, Michael Roth, Annie Louis, Nathanael Chambers, and James Allen.
\newblock Lsdsem 2017 shared task: The story cloze test.
\newblock In {\em Proceedings of the 2nd Workshop on Linking Models of Lexical, Sentential and Discourse-level Semantics}, pages 46--51, 2017.

\bibitem{nagel2020up}
Markus Nagel, Rana~Ali Amjad, Mart Van~Baalen, Christos Louizos, and Tijmen Blankevoort.
\newblock Up or down? adaptive rounding for post-training quantization.
\newblock In {\em International Conference on Machine Learning}, pages 7197--7206. PMLR, 2020.

\bibitem{narayan2018don}
Sameer Narayan, Andre Martins, Alessandro Sordoni, Philip Bachman, Aaron Courville, and Yoshua Bengio.
\newblock Don't give me the details, just the summary!: topic-aware convolutional neural networks for extreme summarization.
\newblock In {\em Proceedings of the 2018 Conference on Empirical Methods in Natural Language Processing}, pages 3706--3716, 2018.

\bibitem{cuBLAS}
NVIDIA.
\newblock cublas.
\newblock \url{"https://developer.nvidia.com/cublas"}, 2023.

\bibitem{TensorRT-LLM}
NVIDIA.
\newblock Tensorrt-llm.
\newblock \url{"https://github.com/NVIDIA/TensorRT-LLM/"}, 2023.

\bibitem{park2022nuqmm}
Gunho Park, Baeseong Park, Se~Jung Kwon, Byeongwook Kim, Youngjoo Lee, and Dongsoo Lee.
\newblock nuqmm: Quantized matmul for efficient inference of large-scale generative language models.
\newblock {\em arXiv preprint arXiv:2206.09557}, 2022.

\bibitem{raffel2020exploring}
Colin Raffel, Noam Shazeer, Adam Roberts, Katherine Lee, Sharan Narang, Michael Matena, Yanqi Zhou, Wei Li, and Peter~J Liu.
\newblock Exploring the limits of transfer learning with a unified text-to-text transformer.
\newblock {\em The Journal of Machine Learning Research}, 21(1):5485--5551, 2020.

\bibitem{scao2022bloom}
Teven~Le Scao, Angela Fan, Christopher Akiki, Ellie Pavlick, Suzana Ili{\'c}, Daniel Hesslow, Roman Castagn{\'e}, Alexandra~Sasha Luccioni, Fran{\c{c}}ois Yvon, Matthias Gall{\'e}, et~al.
\newblock Bloom: A 176b-parameter open-access multilingual language model.
\newblock {\em arXiv preprint arXiv:2211.05100}, 2022.

\bibitem{shang2023pb}
Yuzhang Shang, Zhihang Yuan, Qiang Wu, and Zhen Dong.
\newblock Pb-llm: Partially binarized large language models.
\newblock {\em arXiv preprint arXiv:2310.00034}, 2023.

\bibitem{shen2020q}
Sheng Shen, Zhen Dong, Jiayu Ye, Linjian Ma, Zhewei Yao, Amir Gholami, Michael~W Mahoney, and Kurt Keutzer.
\newblock {Q-BERT}: Hessian based ultra low precision quantization of bert.
\newblock In {\em AAAI}, pages 8815--8821, 2020.

\bibitem{tao2022compression}
Chaofan Tao, Lu~Hou, Wei Zhang, Lifeng Shang, Xin Jiang, Qun Liu, Ping Luo, and Ngai Wong.
\newblock Compression of generative pre-trained language models via quantization.
\newblock {\em arXiv preprint arXiv:2203.10705}, 2022.

\bibitem{tata2003piqa}
Sandeep Tata and Jignesh~M Patel.
\newblock Piqa: An algebra for querying protein data sets.
\newblock In {\em 15th International Conference on Scientific and Statistical Database Management, 2003.}, pages 141--150. IEEE, 2003.

\bibitem{touvron2023llama}
Hugo Touvron, Thibaut Lavril, Gautier Izacard, Xavier Martinet, Marie-Anne Lachaux, Timoth{\'e}e Lacroix, Baptiste Rozi{\`e}re, Naman Goyal, Eric Hambro, Faisal Azhar, et~al.
\newblock Llama: Open and efficient foundation language models.
\newblock {\em arXiv preprint arXiv:2302.13971}, 2023.

\bibitem{van2023fp8}
Mart van Baalen, Andrey Kuzmin, Suparna~S Nair, Yuwei Ren, Eric Mahurin, Chirag Patel, Sundar Subramanian, Sanghyuk Lee, Markus Nagel, Joseph Soriaga, et~al.
\newblock Fp8 versus int8 for efficient deep learning inference.
\newblock {\em arXiv preprint arXiv:2303.17951}, 2023.

\bibitem{wei2023outlier}
Xiuying Wei, Yunchen Zhang, Yuhang Li, Xiangguo Zhang, Ruihao Gong, Jinyang Guo, and Xianglong Liu.
\newblock Outlier suppression+: Accurate quantization of large language models by equivalent and optimal shifting and scaling.
\newblock {\em arXiv preprint arXiv:2304.09145}, 2023.

\bibitem{williams2023does}
Miles Williams and Nikolaos Aletras.
\newblock How does calibration data affect the post-training pruning and quantization of large language models?
\newblock {\em arXiv preprint arXiv:2311.09755}, 2023.

\bibitem{wu2023understanding}
Xiaoxia Wu, Cheng Li, Reza~Yazdani Aminabadi, Zhewei Yao, and Yuxiong He.
\newblock Understanding int4 quantization for transformer models: Latency speedup, composability, and failure cases.
\newblock {\em arXiv preprint arXiv:2301.12017}, 2023.

\bibitem{wu2023zeroquant}
Xiaoxia Wu, Zhewei Yao, and Yuxiong He.
\newblock Zeroquant-fp: A leap forward in llms post-training w4a8 quantization using floating-point formats.
\newblock {\em arXiv preprint arXiv:2307.09782}, 2023.

\bibitem{wu2022extreme}
Xiaoxia Wu, Zhewei Yao, Minjia Zhang, Conglong Li, and Yuxiong He.
\newblock Extreme compression for pre-trained transformers made simple and efficient.
\newblock {\em arXiv preprint arXiv:2206.01859}, 2022.

\bibitem{xia2023flash}
Haojun Xia, Zhen Zheng, Yuchao Li, Donglin Zhuang, Zhongzhu Zhou, Xiafei Qiu, Yong Li, Wei Lin, and Shuaiwen~Leon Song.
\newblock Flash-llm: Enabling cost-effective and highly-efficient large generative model inference with unstructured sparsity.
\newblock {\em arXiv preprint arXiv:2309.10285}, 2023.

\bibitem{xia2023sheared}
Mengzhou Xia, Tianyu Gao, Zhiyuan Zeng, and Danqi Chen.
\newblock Sheared llama: Accelerating language model pre-training via structured pruning.
\newblock {\em arXiv preprint arXiv:2310.06694}, 2023.

\bibitem{xiao2022smoothquant}
Guangxuan Xiao, Ji~Lin, Mickael Seznec, Julien Demouth, and Song Han.
\newblock Smoothquant: Accurate and efficient post-training quantization for large language models.
\newblock {\em arXiv preprint arXiv:2211.10438}, 2022.

\bibitem{yao2023zeroquant-hero}
Zhewei Yao, Reza~Yazdani Aminabadi, Stephen Youn, Xiaoxia Wu, Elton Zheng, and Yuxiong He.
\newblock Zeroquant-hero: Hardware-enhanced robust optimized post-training quantization framework for w8a8 transformers.
\newblock {\em arXiv preprint arXiv:2310.17723}, 2023.

\bibitem{yao2022zeroquant}
Zhewei Yao, Reza~Yazdani Aminabadi, Minjia Zhang, Xiaoxia Wu, Conglong Li, and Yuxiong He.
\newblock Zeroquant: Efficient and affordable post-training quantization for large-scale transformers.
\newblock {\em arXiv preprint arXiv:2206.01861}, 2022.

\bibitem{yao2023zeroquant}
Zhewei Yao, Xiaoxia Wu, Cheng Li, Stephen Youn, and Yuxiong He.
\newblock Zeroquant-v2: Exploring post-training quantization in llms from comprehensive study to low rank compensation.
\newblock {\em arXiv preprint arXiv:2303.08302}, 2023.

\bibitem{yuan2023rptq}
Zhihang Yuan, Lin Niu, Jiawei Liu, Wenyu Liu, Xinggang Wang, Yuzhang Shang, Guangyu Sun, Qiang Wu, Jiaxiang Wu, and Bingzhe Wu.
\newblock Rptq: Reorder-based post-training quantization for large language models.
\newblock {\em arXiv preprint arXiv:2304.01089}, 2023.

\bibitem{zafrir2019q8bert}
Ofir Zafrir, Guy Boudoukh, Peter Izsak, and Moshe Wasserblat.
\newblock {Q8BERT}: Quantized 8bit bert.
\newblock {\em arXiv preprint arXiv:1910.06188}, 2019.

\bibitem{zhang2022opt}
Susan Zhang, Stephen Roller, Naman Goyal, Mikel Artetxe, Moya Chen, Shuohui Chen, Christopher Dewan, Mona Diab, Xian Li, Xi~Victoria Lin, et~al.
\newblock Opt: Open pre-trained transformer language models.
\newblock {\em arXiv preprint arXiv:2205.01068}, 2022.

\bibitem{zhang2019qpytorch}
Tianyi Zhang, Zhiqiu Lin, Guandao Yang, and Christopher~De Sa.
\newblock Qpytorch: A low-precision arithmetic simulation framework, 2019.

\bibitem{zhang2023integer}
Yijia Zhang, Lingran Zhao, Shijie Cao, Wenqiang Wang, Ting Cao, Fan Yang, Mao Yang, Shanghang Zhang, and Ningyi Xu.
\newblock Integer or floating point? new outlooks for low-bit quantization on large language models.
\newblock {\em arXiv preprint arXiv:2305.12356}, 2023.

\bibitem{zheng2023judging}
Lianmin Zheng, Wei-Lin Chiang, Ying Sheng, Siyuan Zhuang, Zhanghao Wu, Yonghao Zhuang, Zi~Lin, Zhuohan Li, Dacheng Li, Eric.~P Xing, Hao Zhang, Joseph~E. Gonzalez, and Ion Stoica.
\newblock Judging llm-as-a-judge with mt-bench and chatbot arena, 2023.

\bibitem{zheng2023codegeex}
Qinkai Zheng, Xiao Xia, Xu~Zou, Yuxiao Dong, Shan Wang, Yufei Xue, Zihan Wang, Lei Shen, Andi Wang, Yang Li, Teng Su, Zhilin Yang, and Jie Tang.
\newblock Codegeex: A pre-trained model for code generation with multilingual evaluations on humaneval-x.
\newblock In {\em KDD}, 2023.

\end{thebibliography}
}

% \clearpage
% \onecolumn
\appendix

%%%%%%%%%%%%%%%%%%%
% Re-count the Figure/Algorithm/Tables after this point. 
%%%%%%%%%%%%%%%%%%%
\counterwithin{figure}{section}
\counterwithin{table}{section}

% \section{Other Efficient Training Approaches}
% \label{sec:other_efficient_training_approaches}
\section{Background of Quantization}
\label{sec:background_of_quantization}
% Note that the following paragraph directly quoted from \cite{yao2022zeroquant} as this is a very well-standadized method.
% Quantization maps floating point (e.g., FP16/FP32) numbers to integer numbers (e.g., INT4/INT8) so that lower memory usage (weight quantization) and faster integer arithmetic (weight-and-activation quantization) can be achieved compared to the floating point format. 
% In this work, we are focusing on uniform quantization, i.e., 
% \begin{equation}
% \small
% \label{eq:quantization_formula}
% Q(x) = \text{INT}\big({(x-Z)}/{S}\big)-Z,
% \end{equation}
% where $Q$ is the quantization function, $x$ is a floating point input vector/tensor, $S$ is a real valued scaling factor, and $Z$ is an integer zero point. 
% Based on different settings, the quantization method can be viewed as (1) symmetric vs. asymmetric quantization ($Z=0$ or not), (2) fine-grained vs. coarse-grained quantization (how to partition the input x and get its associated scaling factor, e.g., matrix wise or row wise). 

Throughout this work, we focus on post-training quantization (\ptq), i.e., no or minimal training effort is applied after quantization, for which large accuracy degradation usually exhibits for coarse-grained quantization (per matrix/tensor) due to their large quantization error.
Particularly, we use the per-row quantization (one row of the weight matrix) from~\cite{yao2022zeroquant} as our coarsest-grained quantization method, 
and we use block-k quantization (for every k elements, they have their own scaling factor and/or zero point) as our finer-grained quantization scheme.

\begin{table}[H]
\caption{GPT4-evaluation for the same model with different precision \cite{zheng2023judging}. There is no clear relation between different bits and rating performance.} \label{table:fastchat}
\begin{adjustbox}{width=0.999\linewidth}
\begin{tabular}{lcccccccc}\toprule 
    Model (Precision)               & Writing & Roleplay & Coding & STEM & Humanities & Reasoning & Math & Extraction \\ \midrule
vicuna-7b-v1.5 (Baseline)& 7.55    & 7.95     & 3.21   & 8.36 & 9.68       & 4.6       & 3.6  & 6.4        \\
vicuna-7b-v1.5 (INT4)  & 7.37    & 7.60     & 3.21   & 8.61 & 9.34       & 5.7       & 2.6  & 6.2        \\
vicuna-7b-v1.5 (FP6)  & 7.55    & 7.88     & 3.67   & 8.55 & 9.21       & 4.8       & 2.2  & 6.1        \\\midrule
vicuna-13b-v1.5 (Baseline) & 7.90    & 7.75     & 2.73   & 7.32 & 9.24       & 5.2       & 2.2  & 6.0        \\
vicuna-13b-v1.5 (INT4)  & 8.42    & 7.56     & 2.91   & 7.93 & 9.58       & 4.7       & 2.7  & 6.4        \\
vicuna-13b-v1.5 (FP6) & 8.06    & 7.60     & 2.94   & 7.76 & 9.31       & 4.8       & 2.8  & 5.9       \\\bottomrule
\end{tabular}
\end{adjustbox}
\end{table}

 \begin{table}[H]
 \caption{\textbf{Zero-Shot Evaluation (Accuracy)}. Compare between GPTQ$_{C4}$ and RTN quantization algorithms for INT4 weight (W4A16) on \llama-1B (Top) and \llama-13B (Bottom). We apply fine-grain quantization (FGQ) in which the block-size is 256 elements per scale except for \llama-1B's (which is 128). arcC (arcE) stands for arc\_challenges (arc\_easy).
 }
  \begin{adjustbox}{width=0.999\linewidth}\label{table:acc-int-all-app}
 \begin{tabular}{llllllllllllll}\toprule
        Models             &   Precision   &   FGQ    &  arcC & arcE & boolq & cb    & copa  & piqa  & rte   & wic   & wsc   & storycloze & MEAN \\\midrule 
\multirow{6}{*}{\llama-1B }&  FP16 &  N/A &  26.71 & 53.11 & 61.13 & 39.29 & 76.00    & 73.83 & 50.18 & 50.00    & 36.54 & 69.64 & 53.64   \\\cdashline{2-14}
\multirow{6}{*}{(4096-seq)}&  INT4-GPTQ    & \cmark &  {26.37}          & 50.59     &  {61.59} & 46.43 &  {79.00}    & {73.34} & 48.01 & 50.00    & 36.54 & 68.24      & {54.01} \\
                     &  INT4-RTN & \cmark &  26.11          &  {51.09}     & 58.07 &  {50.00}    & 74.00    & 72.91 & {48.38} & 50.00    & 36.54 & {68.36}      & 53.55 \\\cdashline{2-14}
                                           &  FP5-GPTQ &  \xmark &25.51 & 51.85 & 61.22 & 32.14 & 80.00 & 73.01 & 48.38 &50.00& 36.54 & 68.94 & 52.76\\
                     & FP5-RTN  & \xmark & 26.62 & 50.67 & 61.41 & 39.29 & 79.00 & 72.85 & 50.18 &50.00& 36.54 & 69.19 & 53.58 \\
                                  &  FP6-GPTQ &  \xmark &26.37 & 52.53 & 61.19 & 42.86 & 75.00     & 73.5  & 51.26 &50.00   & 36.54 & 69.76 & 53.90 \\
                     & FP6-RTN  & \xmark & 26.37 & 52.95 & 60.95 & 37.50  & 78.00    & 73.39 & 54.51 &50.00   & 36.54 & 69.64 & 53.99\\\midrule
\multirow{6}{*}{\llama-13B}& FP16& N/A&  43.86 & 74.58 & 68.53 & 50.00 & 90.00 & 79.00 & 65.34 & 50.00 & 35.58 & 78.23 & 63.51   \\\cdashline{2-14}
\multirow{6}{*}{(2048-seq)}&  INT4-GPTQ    & \cmark &  43.00          & 73.44     & {67.83} & 41.07 & {93.00} & 78.78 & 62.45 & {50.16} & 36.54 & 78.17      & 62.44 \\
                     & INT4-RTN& \cmark &  {44.03}          & {74.45}     & 67.37 & {44.64} & 91.00 & {78.84} & {63.18} & 49.84 & 36.54 & {78.42}      & {62.83}\\\cdashline{2-14}
                     &  FP5-GPTQ &  \xmark &42.92 & 73.70  & 65.81 & 44.64 & 90.00 & 78.67 & 64.62 & 50.00    & 36.54 & 78.23 & 62.51\\
                     & FP5-RTN  & \xmark & 41.72 & 74.03 & 68.47 & 39.29 & 90.00 & 78.56 & 62.45 & 50.31 & 36.54 & 78.61 & 62.00  \\
                      &  FP6-GPTQ &  \xmark &43.69 & 73.99 & 67.28 & 48.21 & 90.00 & 78.84 & 64.98 & 50.31 & 36.54 & 78.42 & 63.23 \\
                     & FP6-RTN  & \xmark & 43.77 & 74.20  & 68.38 & 46.43 & 91.00    & 78.84 & 65.34 & 49.84 & 36.54 & 78.23 & 63.26 \\
                     \midrule
    \multirow{6}{*}{\llama-65B}& FP16 &  N/A &  47.01 & 75.08 & 82.32 & 64.29 & 91.00 & 81.61 & 71.48 & 58.31 & 60.58 & 79.57 & 71.13  \\
    
    \cdashline{2-14}
\multirow{6}{*}{(2048-seq)}&  INT4-GPTQ  & \cmark & 46.84 & 75.08 & 80.76 & 58.93 & 94.00 & 81.18 &  {72.92} & 56.27 & 60.58 & 79.31 & 70.59  \\
                     & INT4-RTN  & \cmark &  {47.10} &  {75.25} &  {81.47} &  {62.50} &  {95.00} &  {81.23} & 69.68 &  {57.21} &  {62.50} &  {79.63} & {71.16} \\\cdashline{2-14}
                    &  FP5-GPTQ &  \xmark & 46.50 & 75.51 & 82.35 & 69.64 & 93.00 & 81.28 & 71.84 & 57.05 & 57.69 & 79.76 & 71.46  \\
                     & FP5-RTN  & \xmark & 46.50 & 75.59 & 82.87 & 60.71 & 94.00 & 81.39 & 73.65 & 57.21 & 60.58 & 80.08 & 71.26\\
                     &  FP6-GPTQ &  \xmark & 46.84 & 74.96 & 82.51 & 64.29 & 91.00 & 81.23 & 70.04 & 59.72 & 61.54 & 79.63 & 71.18 \\
                     & FP6-RTN  & \xmark & 47.10  & 74.66 & 82.69 & 64.29 & 92.00 & 81.99 & 70.76 & 58.15 & 57.69 & 79.31 & 70.86\\
\bottomrule  
\end{tabular}
\end{adjustbox}
 \end{table}

\section{Detailed Performance of GPU Kernels}
\label{sec:DetailedKernelLatency}

The shapes of weight matrices are set according to the model specification of \llama-1B, \llama-13B, and \llama-65B, respectively.
We mainly evaluate the kernel performance when the inference batch size is 8.
As for the Fine-grained INT4 kernel, we set its quantization group size to 128 for the best of its performance.
All the kernel latency shown here is measured in milliseconds.

\begin{table}[H]
\caption{The corresponding number for \fref{fig:KernelPerformance}.}\label{table:kernel}
\begin{adjustbox}{width=0.999\linewidth}
\begin{tabular}{lcccccccc}\toprule 
    Layer Name  & Weight Shape  & Input Shape   & cuBLAS    & Fine-grained INT4 & FP6 (w/o Bias-Shift)  & FP6 (w/ Bias-Shift)   \\ \midrule
    FFN1-1b     & 5504*2048     & 2048*8        & 0.016     & 0.019             & 0.021                 & 0.013                 \\
    FFN2-1b     & 2048*5504     & 5504*8        & 0.02      & 0.043             & 0.022                 & 0.016                 \\
    FFN1-13b    & 13824*5120    & 5120*8        & 0.118     & 0.044             & 0.063                 & 0.052                 \\
    FFN2-13b    & 5120*13824    & 13824*8       & 0.109     & 0.106             & 0.068                 & 0.053                 \\
    FFN1-65b    & 22016*8192    & 8192*8        & 0.263     & 0.098             & 0.145                 & 0.111                 \\
    FFN2-65b    & 8192*22016    & 22016*8       & 0.266     & 0.167             & 0.157                 & 0.114                 \\ \bottomrule
\end{tabular}
\end{adjustbox}
\end{table}

\end{document}